\PassOptionsToPackage{numbers, compress}{natbib}

\documentclass{article}
\usepackage[preprint]{neurips_2026}

\usepackage[most]{tcolorbox}
\usepackage{fvextra}
\usepackage{caption} 

\usepackage{times}
\usepackage{latexsym}
\usepackage{booktabs,tabularx,array,ragged2e,makecell}
\usepackage[table]{xcolor} 
\usepackage{tcolorbox}
\tcbuselibrary{breakable,skins}
\usepackage{fvextra}   
\usepackage{ragged2e}
\usepackage{array,tabularx} 
\usepackage{caption}

\newcommand{\HL}[2]{%
  \begingroup
  \setlength{\fboxsep}{1.2pt}%
  \colorbox{#1!30}{%
    \parbox[t]{\dimexpr\linewidth-2\fboxsep\relax}{%
      \ttfamily\raggedright #2%
    }%
  }%
  \endgroup
}
\newcommand{\HLcyan}[1]{\HL{cyan}{#1}}
\newcommand{\HLred}[1]{\HL{red}{#1}}
\newcommand{\HLorange}[1]{\HL{orange}{#1}}
\newcommand{\HLgreen}[1]{\HL{green}{#1}}

\definecolor{hlblue}{RGB}{220,235,255}
\definecolor{hlframe}{RGB}{120,160,220}

\newtcolorbox{highlightbox}{
  enhanced,
  breakable,
  colback=hlblue,
  colframe=hlframe,
  boxrule=0.6pt,
  arc=1.5mm,
  left=1.2mm,right=1.2mm,top=0.8mm,bottom=0.8mm,
}

\newcolumntype{Q}{>{\RaggedRight\arraybackslash}X}
\newcolumntype{W}{>{\ttfamily\RaggedRight\arraybackslash}X}
\newcolumntype{J}[1]{>{\RaggedRight\arraybackslash}p{#1}}
\newcolumntype{K}[1]{>{\centering\arraybackslash}p{#1}}

\usepackage{booktabs}
\usepackage{ragged2e}
\usepackage{tcolorbox}
\tcbuselibrary{skins,breakable}

\usepackage{fancyvrb}
\usepackage{fvextra} 
\usepackage{amsmath}
\DefineVerbatimEnvironment{DAGPlanVrb}{Verbatim}{
  breaklines=true,
  breakanywhere=true,
  breakautoindent=true
}

\newlength{\DAGLeftW}
\newlength{\DAGRightW}
\newcommand{\DAGSep}{\par\smallskip\hrule height 0.4pt\par\smallskip}

\usepackage{algorithm}
\usepackage{algpseudocode} 
\usepackage{array} 
\usepackage{enumitem}
\usepackage{amssymb}
\usepackage{booktabs}
\usepackage{tabularx}
\usepackage{array}
\usepackage{ragged2e}
\usepackage{url} 
\usepackage{graphicx}
\usepackage{tcolorbox}
\tcbuselibrary{breakable,skins}
\usepackage{fvextra} 
\usetikzlibrary{arrows.meta,positioning}

\tikzset{
  dagbox/.style={
    draw,
    minimum width=10mm,
    minimum height=6mm,
    inner sep=1pt,
    font=\scriptsize
  },
  dagarrow/.style={-Latex, line width=0.35pt}
}
\newcommand{\MiniDAGGraphA}{%
\begin{tikzpicture}[baseline=(current bounding box.north), node distance=4mm and 8mm]
  \node[dagbox] (s1) {S1};
  \node[dagbox, right=8mm of s1] (s2) {S2};
  \node[dagbox, below=4mm of s1] (s3) {S3};
  \node[dagbox, below=4mm of s3] (s4) {S4};

  \draw[dagarrow] (s1) -- (s2);
  \draw[dagarrow] (s1) -- (s3);
  \draw[dagarrow] (s2) -- (s3);
  \draw[dagarrow] (s3) -- (s4);
\end{tikzpicture}%
}

\newcommand{\MiniDAGGraphB}{%
\begin{tikzpicture}[baseline=(current bounding box.north), node distance=4mm and 8mm]
  \node[dagbox] (s1) {S1};
  \node[dagbox, below=4mm of s1] (s2) {S2};

  \node[dagbox, below left=6mm and 6mm of s2] (s3) {S3};
  \node[dagbox, below right=6mm and 6mm of s2] (s4) {S4};

  \draw[dagarrow] (s1) -- (s2);
  \draw[dagarrow] (s2) -- (s3);
  \draw[dagarrow] (s2) -- (s4);
  \draw[dagarrow] (s3) -- (s4);
\end{tikzpicture}%
}

\newcolumntype{L}[1]{>{\RaggedRight\arraybackslash}p{#1}}
\newcolumntype{C}[1]{>{\Centering\arraybackslash}p{#1}}
\usepackage{booktabs}
\usepackage{siunitx}
\usepackage{graphicx}  
\usepackage[table]{xcolor} 

\usepackage{listings}

\setlist[itemize]{leftmargin=*, topsep=2pt, itemsep=2pt, parsep=0pt}

\usepackage[T1]{fontenc}

\usepackage[utf8]{inputenc}

\usepackage{microtype}

\usepackage{inconsolata}

\usepackage{graphicx}

\title{SPIN: Structural LLM Planning via Iterative Navigation for Industrial Tasks}

\author{
  Yusuke Ozaki \\
  University at Albany \\
  Kwansei Gakuin University \\
  \texttt{inq31294@kwansei.ac.jp}
  \And
  Dhaval Patel \\
  IBM, New York \\
  \texttt{pateldha@us.ibm.com}
}

\begin{document}

\maketitle

\begin{abstract}
Industrial LLM agent systems often separate planning from execution, yet LLM planners frequently produce structurally invalid or unnecessarily long workflows, leading to brittle failures and avoidable tool and API cost. We propose \texttt{SPIN}, a planning wrapper that combines validated Directed Acyclic Graph (DAG) planning with prefix based execution control. \texttt{SPIN} enforces a strict DAG contract through \texttt{\_validate\_plan\_text} and repair prompting, producing executable plans before downstream execution, and then evaluates DAG prefixes incrementally to stop when the current prefix is sufficient to answer the query. On AssetOpsBench, across 261 scenarios, \texttt{SPIN} reduces executed tasks from 1061 to 623 and improves \emph{Accomplished} from 0.638 to 0.706, while reducing tool calls from 11.81 to 6.82 per run. On MCP Bench, the same wrapper improves planning, grounding, and dependency related scores for both GPT OSS1 and Llama 4 Maverick.
\end{abstract}

\section{Introduction}
\label{sec:introduction}

Industrial asset operations and maintenance workflows are inherently multi-step, cross-functional, and constrained, requiring decisions that are not only correct but also auditable and robust to heterogeneous tool latencies and failures.
In benchmarks such as AssetOpsBench, the planner's output serves as an interface contract for downstream executors.
A key practical challenge is that small structural errors in a plan, such as incorrect indexing, dependency references, or tool and agent names, can cause hard execution failures and obscure whether the true problem is task reasoning or interface brittleness.
Moreover, executing long, conservative workflows end to end can be inefficient when external actions dominate cost and runtime, even though partial progress may already be sufficient to answer the user.
These challenges motivate methods that (i) make the planner--executor boundary explicit and measurable, and (ii) support cost-aware execution policies that terminate when further actions are unlikely to change the final outcome.

Prior work on LLM agents has largely focused on improving action generation, for example through interleaving reasoning with tool use \citep{yao2023react,singh2025artist} or revising outputs through reflection and self-feedback \citep{shinn2023reflexion,liu-etal-2025-instruct,yang-etal-2025-confidence}.
However, the planner--executor interface is often left implicit, so downstream evaluation can be dominated by brittle formatting errors rather than true task difficulty, especially in domain-grounded benchmarks \citep{patel2025assetopsbench}.
Recent evidence further suggests that LLMs cannot be relied upon to consistently verify and repair planning outputs without external checks \citep{stechly2025selfverification,gou2023critic,chang2025sagallm}.
Although structured generation can improve schema compliance \citep{geng2025jsonschemabench,lu-etal-2025-learning} and search-based methods can use external tests to guide generation \citep{katz2024thoughtofsearch,cao2024autotos}, these ideas are rarely combined with a dependency-consistent task graph representation and a cost-sensitive execution policy such as early stopping in a domain-grounded agent pipeline.

\begin{figure*}[t!]
  \centering
\includegraphics[width=\textwidth]{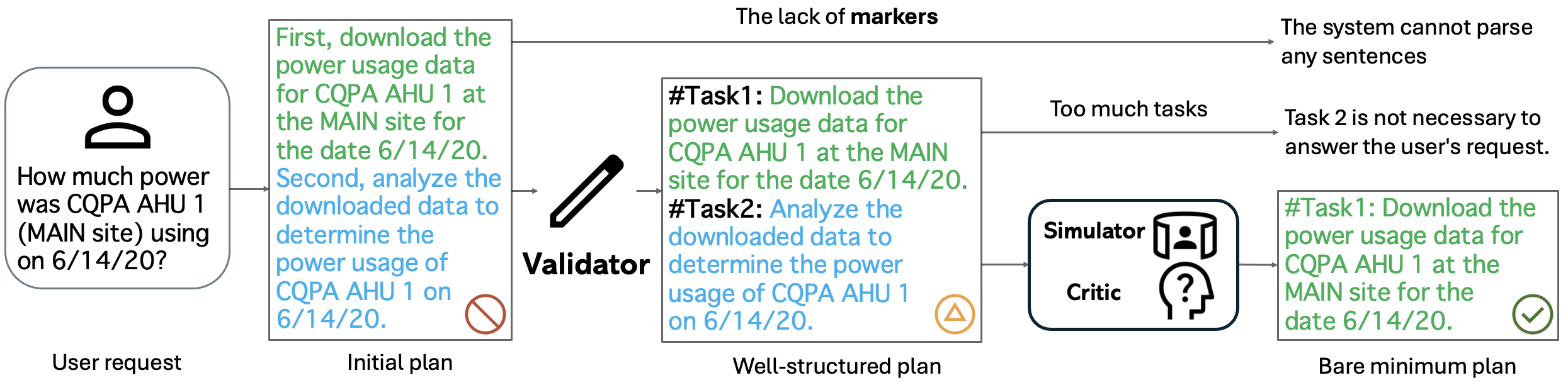}
  \caption{Working example of validation and SPIN system.}
  \label{fig:working_example}
\end{figure*}

We present \texttt{SPIN}, a planning wrapper for tool using LLM agents that makes plan structure an explicit executable interface, instantiated on AssetOpsBench and additionally evaluated on MCP Bench \citep{wang2025mcpbenchbenchmarkingtoolusingllm}. Starting from a planner generated \texttt{plan\_text}, \texttt{SPIN} validates and repairs it into a dependency aware DAG serialization that an executor can consume deterministically. For execution, \texttt{SPIN} applies a simulator and critic based prefix policy, adapted from SPIRAL \citep{zhang2025spiralsymbolicllmplanning}, to stop once the current DAG prefix is sufficient to answer the query. 

\paragraph{Contributions}
\begin{itemize}
  \item We develop a contract based planning wrapper for tool using LLM agents. Our method combines a validator that enforces a dependency aware DAG format with a prefix evaluation loop in which a simulator and a critic decide whether further execution is necessary.
  \item We present empirical evaluation on two benchmarks with different knowledge settings. The results show improved executability and lower execution burden on AssetOpsBench, and improved planning and grounding related scores on MCP Bench.
  \item We provide a failure mode analysis showing that the gains of \texttt{SPIN} are concentrated in execution behavior, especially reduced repetition and improved state and termination related control. The analysis also suggests that the simulator is a more load bearing component than the critic on the matched scenario set.
\end{itemize}

\section{Related Work}
\label{sec:related_work}

AssetOpsBench is designed for industrial asset operations and maintenance, where operational workflows such as condition monitoring, maintenance planning, and intervention scheduling require coordination across multiple specialized steps, heterogeneous tools, and large operational data sources \citep{patel2025assetopsbench}. This setting differs from many standard agent benchmarks because planner outputs do not remain purely internal reasoning artifacts: they can directly shape downstream execution cost, operational burden, and recovery behavior. In such workflows, the relevant question is not only whether an agent can produce a plausible plan, but also whether that plan can be safely consumed and executed within an execution-heavy industrial pipeline.

A large body of prior work has improved LLM agents through better reasoning, reflection, tool use, or structured generation. Representative examples include interleaving reasoning with acting, verbal reinforcement learning and reflection, dynamic reflection control, decomposed self-correction, reinforcement learning for tool-integrated agents, and tool-interactive critiquing \citep{yao2023react,shinn2023reflexion,liu-etal-2025-instruct,yang-etal-2025-confidence,singh2025artist,gou2023critic}. Related systems work has also emphasized stronger validation and consistency guarantees in multi-agent planning pipelines \citep{chang2025sagallm}. In parallel, another line of work studies grammar-constrained decoding, schema-constrained outputs, benchmarked structured generation, and schema-level reinforcement learning \citep{geng-etal-2023-grammar,geng2025jsonschemabench,lu-etal-2025-learning,openai2024structuredoutputs}. These directions are highly relevant to our setting, but our emphasis is different: when agent systems are deployed through external tools in industrial applications, the output must be not only well formed, but also machine-consumable as an execution interface. Accordingly, we treat planner outputs as executable DAG interfaces rather than as generic structured text.

A complementary line of work evaluates LLM agents on realistic, multi-tool environments via the Model Context Protocol. MCP-Bench probes agent behavior across a large suite of MCP servers and heterogeneous real-world tools, measuring tool selection, planning, grounding, and parameter accuracy under task specifications that require multi-hop, cross-domain tool use \citep{wang2025mcpbenchbenchmarkingtoolusingllm}. Our work shares the view that agents must be evaluated against execution-faithful tool interfaces rather than synthetic APIs, but targets a different regime: industrial asset-operations workflows in which planner outputs are consumed downstream as executable DAGs, and where structural validity and execution cost are first-class evaluation criteria.

A third line of work improves planning through search, tests, simulator-guided reasoning, or more efficient reasoning--action organization \citep{xu2023rewoo,katz2024thoughtofsearch,cao2024autotos,zhang2025spiralsymbolicllmplanning}. ReWOO reduces redundant interleaving by decoupling reasoning from observations, Thought of Search and AutoToS emphasize efficient and correct search components, and SPIRAL uses a planner--simulator--critic decomposition to improve search over predicted outcomes. Our contribution is closest to this family, but differs in optimization target. We do not primarily aim to improve search quality over imagined futures. Instead, we focus on industrial settings in which downstream execution can dominate practical cost: as language-model inference becomes cheaper and faster over time, the dominant burden may shift toward executable steps, tool invocations, data transfer, API traffic, and recovery-inducing failures, especially in large-scale data-intensive workflows. This execution-sensitive perspective is also consistent with recent calls for agent evaluation beyond raw accuracy alone \citep{kapoor2024aiagentsmatter}. This is why we treat structural validity and prefix-based control as first-class concerns, and evaluate planner outputs with execution-sensitive criteria in AssetOpsBench.

\section{Method}
\label{sec:method}

\begin{figure}[h!]
  \centering
\includegraphics[width=0.7\linewidth]{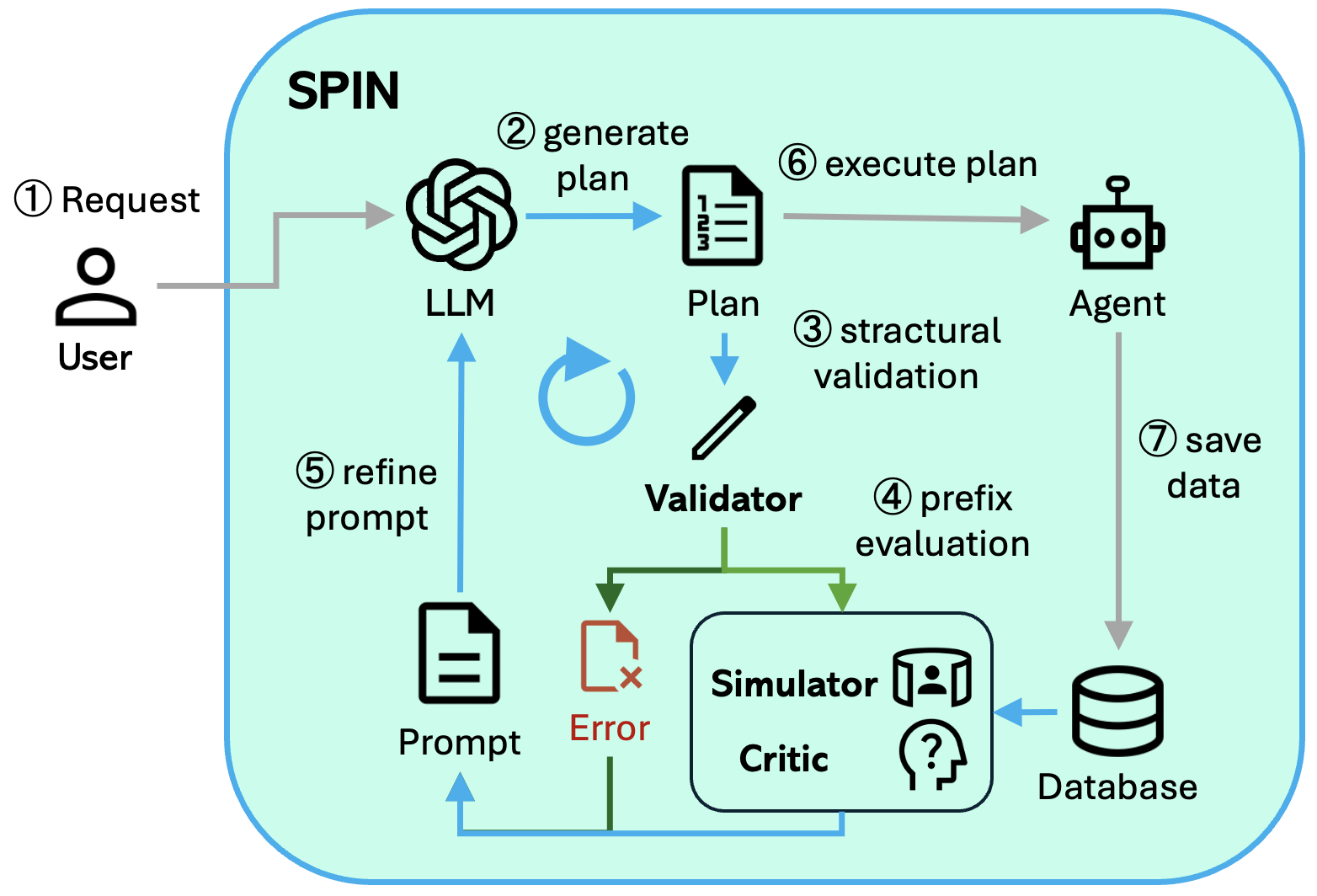}
  \caption{System overview of validated DAG planning and prefix evaluation  by Simulator and Critic.}
  \vspace{-0.2in}
  \label{fig:system_overview}
\end{figure}

We build a planning and execution workflow for tool using LLM agents, instantiated on AssetOpsBench and later evaluated on MCP Bench, where a planner first produces a task graph (Table~\ref{tab:dag_examples}) and an evaluation loop determines how much of the plan is necessary to answer the user query (See Figure~\ref{fig:working_example}, \ref{fig:dag_visual}). Our pipeline has two core components (Figure~\ref{fig:system_overview}): (1) a strict plan validator/repair loop that enforces a machine readable DAG format, and (2) a simulator and critic evaluation loop with early stopping. Our prefix evaluation loop is inspired by the simulator critic decomposition in SPIRAL, but differs in that we operate over validated DAG prefixes and do not use Monte Carlo Tree Search style reflective search.

\subsection{Formal Problem Setup}
\label{sec:formal_setup}

We study a tool-using planning setting in which a planner receives a user query
$q$ and produces a textual multi-step workflow $x$ that must be consumed by a
downstream executor. We represent a parsed plan as an ordered set of nodes
$p = \{n_1, \dots, n_N\}$, where each node
$n_i = (t_i, a_i, d_i, e_i)$
contains a task description $t_i$, an assigned agent $a_i$, a dependency set
$d_i \subseteq \{1,\dots,i-1\}$, and an expected output specification $e_i$.
In our implementation, the raw plan text is serialized as four aligned lists:
\texttt{\#TaskN}, \texttt{\#AgentN}, \texttt{\#DependencyN}, and
\texttt{\#ExpectedOutputN}. 

We treat plan executability as structural validity of this serialized interface
prior to downstream execution. Let $V(x)$ denote a validator that returns a
validity indicator and an error set,
$V(x) = (\mathrm{valid}(x), E(x))$.
A plan is considered valid if it satisfies at least the following constraints:
(i) the four lists have matching lengths, (ii) indices are consecutive from
$1$ to $N$, (iii) each dependency refers only to a valid previous node, and
(iv) each assigned agent belongs to a predefined admissible set. These
constraints define the executable plan interface targeted by our repair loop. 

Given a valid plan $p$, we evaluate increasingly longer prefixes
$p^{(k)} = \{n_1,\dots,n_k\}$ rather than executing the entire workflow by
default. For each prefix, a simulator $S$ predicts the candidate outcome of the
current execution state, and a critic $C$ returns a structured judgment
containing a status, a Boolean \texttt{can\_answer\_now} flag, and a rationale.
This defines a stopping policy over plan prefixes:
the system halts once the current prefix is judged sufficient to answer the
original query $q$. 

Our method therefore optimizes two related objectives. The first is
\emph{executability}: increasing the probability that planner outputs are
machine-consumable before execution. The second is a
\emph{quality--effort trade-off}: reducing external execution effort, such as
executed tasks, tool calls, API calls, and elapsed time, while preserving
sufficient answer quality. Importantly, this objective does not require all
forms of cost to decrease simultaneously; in particular, additional simulator
and critic prompting may increase internal token usage even when external
execution effort decreases. 

\subsection{Executable DAG Validation and Repair}
\label{sec:validation_repair}

We operationalize plan executability as structural validity of the planner output
prior to downstream execution. Rather than directly executing the raw
\texttt{plan\_text}, we first apply a validator that checks alignment across task
fields, index consistency, dependency legality, and agent-name admissibility.
If the plan is invalid, we feed the detected error set back to the planner and
request a repaired plan. This produces a simple error-aware repair loop whose
goal is to transform raw planner outputs into a machine-consumable DAG interface.

\begin{algorithm}[t]
\caption{Validation and Repair Loop}
\label{alg:validation_repair}
\begin{algorithmic}[1]
\Require user query $q$, planner $\Pi$, validator $V$, retry budget $T$
\Ensure valid plan $p$ or failure after budget exhaustion

\State $x \gets \Pi(q)$
\For{$r = 0$ to $T$}
    \State $(\mathrm{valid}, E) \gets V(x)$
    \If{$\mathrm{valid}$}
        \State \Return $x$
    \EndIf
    \If{$r = T$}
        \State \Return \textsc{Fail}
    \EndIf
    \State $x \gets \Pi(q; E)$ \Comment{repair conditioned on detected errors}
\EndFor
\end{algorithmic}
\end{algorithm}

In our implementation, the validator checks whether the four serialized fields
(\texttt{\#TaskN}, \texttt{\#AgentN}, \texttt{\#DependencyN}, and
\texttt{\#ExpectedOutputN}) are aligned, whether indices are consecutive,
whether dependencies refer only to valid previous steps, and whether assigned
agents belong to an admissible set. The output of this procedure is therefore
not a task-level success judgment, but a pre-execution executability judgment
over the planner--executor interface.

\subsection{Prefix Evaluation with Stopping}
\label{sec:prefix_stopping}

Given a validated plan $p = \{n_1,\dots,n_N\}$, we do not execute the full
workflow by default. Instead, we evaluate a sequence of increasingly longer
prefixes
\[
p^{(k)} = \{n_1,\dots,n_k\}, \qquad k=1,\dots,N,
\]
and stop as soon as the current prefix is judged sufficient to answer the
original user query. This defines execution control as a stopping problem over
validated DAG prefixes rather than as unconditional end-to-end execution.

For each prefix $p^{(k)}$, we decompose the decision into two roles. A
\emph{simulator} $S$ predicts the candidate outcome of the current execution
state without requiring the full downstream execution of later nodes. A
\emph{critic} $C$ then evaluates whether the current prefix, together with the
simulated candidate outcome, is already sufficient to answer the query. The
critic returns a structured judgment
\[
C(q, p^{(k)}, \hat{o}_k)
=
(\mathrm{status}_k,\ \mathrm{can\_answer\_now}_k,\ r_k),
\]
where $\hat{o}_k = S(q, p^{(k)})$ is the simulator output,
$\mathrm{status}_k$ is a task-level completion label,
$\mathrm{can\_answer\_now}_k \in \{\mathrm{True}, \mathrm{False}\}$ is a
Boolean sufficiency flag, and $r_k$ is a short rationale.

\begin{algorithm}[t]
\caption{Prefix Evaluation with Stopping}
\label{alg:prefix_stopping}
\begin{algorithmic}[1]
\Require validated plan $p=\{n_1,\dots,n_N\}$, user query $q$, simulator $S$, critic $C$
\Ensure final answer $\hat{y}$, executed prefix length $k^\star$

\For{$k = 1$ to $N$}
    \State construct current prefix $p^{(k)} = \{n_1,\dots,n_k\}$
    \State $\hat{o}_k \gets S(q, p^{(k)})$
    \State $(\mathrm{status}_k, \mathrm{can\_answer\_now}_k, r_k) \gets C(q, p^{(k)}, \hat{o}_k)$
    \If{$\mathrm{can\_answer\_now}_k = \mathrm{True}$ \textbf{and}
        $\mathrm{status}_k \in \{\textsc{Accomplished}, \textsc{PartiallyAccomplished}\}$}
        \State \Return $(\hat{o}_k, k)$
    \EndIf
\EndFor
\State \Return $(\hat{o}_N, N)$
\end{algorithmic}
\end{algorithm}

The stopping decision is therefore given by
\[
\begin{aligned}
\mathrm{Stop}(k)=1
&\iff
\Bigl(\mathrm{status}_k \in
\{\textsc{Accomplished},\textsc{PartiallyAccomplished}\}\Bigr) \\
&\land
\Bigl(\mathrm{can\_answer\_now}_k=\mathrm{True}\Bigr).
\end{aligned}
\]
If this condition is not satisfied, the system extends the prefix and repeats
the same decision process at the next step. If no earlier prefix satisfies the
criterion, the method defaults to the full validated plan.

This procedure induces an explicit quality--effort trade-off. On the one hand,
shorter accepted prefixes reduce \emph{external execution effort}, such as the
number of executed tasks, tool calls, API calls, and elapsed runtime. On the
other hand, the simulator and critic introduce additional \emph{internal
prompting effort}, measured for example in prompt and completion tokens.
Accordingly, SPIN is not designed to reduce every notion of cost
simultaneously. Rather, its objective is to reduce costly downstream execution
while preserving sufficient answer quality through prefix-level stopping.

\section{Experiments}
\label{sec:experiments}

We evaluate three points: (i) whether strict DAG validation and targeted repair improve plan executability before downstream execution, (ii) how prefix evaluation changes outcome quality and execution cost on AssetOpsBench, and (iii) how the same planning wrapper behaves on MCP Bench.

\subsection{Experimental setup and systems}
\label{sec:exp_setup}

On AssetOpsBench, we evaluate 261 scenarios in total, consisting of 141 original in-distribution (IID) scenarios and 120 additional out-of-distribution (OOD) scenarios. The additional OOD scenarios use the same agents and tools as the original benchmark but are instantiated on different assets, so the expanded evaluation should be interpreted not only as a larger in-domain test set but also as a generalization test under asset shift. Each scenario provides a natural language user request and a fixed tool and agent environment including IoT metadata, sensor time series, and work order artifacts. Each scenario provides a natural language user request and a fixed tool and agent environment including IoT metadata, sensor time series, and work order artifacts. We use paired comparisons between the baseline and SPIN based workflow under the same scenario inputs and logging settings. Unless otherwise noted, efficiency metrics are computed over non zero task runs, as reported in Table~\ref{tab:overall_outcome_scale} and Table~\ref{tab:overall_effort}.

\begin{table*}[t]
  \centering
  \scriptsize
  \setlength{\tabcolsep}{1pt}
  \renewcommand{\arraystretch}{1.08}
  \caption{Plan format validation results for \texttt{\_validate\_plan\_text}. Each error cell reports the number of plan files in the directory that triggered the corresponding validation rule at least once, with percentage over 141 total plans per model in that directory. Columns B\#* are \texttt{[BASE]} runs and V\#* are \texttt{[VALID]} runs for each LLM model (Table~\ref{tab:llm_id_to_model}).}
  \label{tab:dag_success_validate_plan_text}

  \begin{tabularx}{\linewidth}{>{\RaggedRight\arraybackslash}X *{10}{c}}
    \toprule
    & \multicolumn{5}{c}{BASE} & \multicolumn{5}{c}{VALID} \\
    \cmidrule(lr){2-6}\cmidrule(lr){7-11}
    Metric & B\#16 & B\#20 & B\#22 & B\#23 & B\#38 & V\#16 & V\#20 & V\#22 & V\#23 & V\#38 \\
    \midrule

    success\_rate & 0.695 & 0.021 & \bfseries 0.979 & 0.745 & 0.943 & \bfseries 1.000 & \bfseries 0.170 & 0.972 & 0.745 & \bfseries 1.000 \\
    ok\_plans     & 98    & 3     & 138   & 105   & 133   & 141   & 24    & 137   & 105   & 141 \\
    \midrule

    \rowcolor[RGB]{190,225,255}
    \texttt{COUNTS\_MISMATCH} & 15 (10.6\%) & 104 (73.8\%) & \bfseries 0 (0.0\%) & \bfseries 0 (0.0\%) & 0 (0.0\%) & \bfseries 0 (0.0\%) & \bfseries 102 (72.3\%) & 4 (2.8\%) & 36 (25.5\%) & 0 (0.0\%) \\

    \rowcolor[RGB]{255,200,200}
    \texttt{AGENT\_UNKNOWN} & 9 (6.4\%) & 123 (87.2\%) & 1 (0.7\%) & 0 (0.0\%) & 7 (5.0\%) & \bfseries 0 (0.0\%) & \bfseries 57 (40.4\%) & \bfseries 0 (0.0\%) & 0 (0.0\%) & \bfseries 0 (0.0\%) \\

    \rowcolor[RGB]{255,220,170}
    \texttt{TASK\_NUMBERS\_NOT\_SEQ} & 33 (23.4\%) & 57 (40.4\%) & \bfseries 0 (0.0\%) & \bfseries 0 (0.0\%) & 0 (0.0\%) & \bfseries 0 (0.0\%) & 57 (40.4\%) & 4 (2.8\%) & 35 (24.8\%) & 0 (0.0\%) \\

    \rowcolor[RGB]{255,220,170}
    \texttt{AGENT\_NUMBERS\_NOT\_SEQ} & 34 (24.1\%) & 66 (46.8\%) & 0 (0.0\%) & 0 (0.0\%) & 1 (0.7\%) & \bfseries 0 (0.0\%) & \bfseries 21 (14.9\%) & 0 (0.0\%) & 0 (0.0\%) & \bfseries 0 (0.0\%) \\

    \rowcolor[RGB]{255,220,170}
    \texttt{DEPENDENCY\_NUMBERS\_NOT\_SEQ} & 33 (23.4\%) & 67 (47.5\%) & 0 (0.0\%) & 0 (0.0\%) & 0 (0.0\%) & \bfseries 0 (0.0\%) & \bfseries 22 (15.6\%) & 0 (0.0\%) & 0 (0.0\%) & 0 (0.0\%) \\

    \rowcolor[RGB]{255,220,170}
    \texttt{EXPECTEDOUTPUT\_NUMBERS\_NOT\_SEQ} & 33 (23.4\%) & 67 (47.5\%) & 0 (0.0\%) & 0 (0.0\%) & 0 (0.0\%) & \bfseries 0 (0.0\%) & \bfseries 21 (14.9\%) & 0 (0.0\%) & 0 (0.0\%) & 0 (0.0\%) \\

    \bottomrule
  \end{tabularx}
\end{table*}

We compare four systems. \textbf{[BASE]} is the original sequential planning and execution pipeline. \textbf{[SPIN]} applies \texttt{\_validate\_plan\_text} to enforce a dependency aware DAG format and then evaluates increasingly longer DAG prefixes, stopping when the critic returns that the current prefix is sufficient to answer the query. \textbf{[SPIN\_wo\_sim]} removes retrieval conditioned simulation from the prefix loop. \textbf{[SPIN\_wo\_cri]} weakens critic based stopping. In addition, we evaluate SPIN on 18 two server grounding tasks from MCP Bench using GPT OSS1 (ID~20) and Llama~4~Maverick (ID~16).

\subsection{AssetOpsBench results}
\label{sec:assetops_results}

Table~\ref{tab:dag_success_validate_plan_text} reports offline plan format validation results. For each planner configuration, we collect raw \texttt{plan\_text} under the baseline prompt (\texttt{B\#*}) and the validator assisted setting (\texttt{V\#*}), and compute (i) \textbf{success\_rate}, the fraction of plans that pass all checks, and (ii) \textbf{error incidence} by rule. The most frequent error types are list-length
inconsistencies
({\setlength{\fboxsep}{1pt}\colorbox[RGB]{190,225,255}{\strut\texttt{COUNTS\_MISMATCH}}}),
unsupported agent names
({\setlength{\fboxsep}{1pt}\colorbox[RGB]{255,200,200}{\strut\texttt{AGENT\_UNKNOWN}}}),
and non-consecutive indices across sections
({\setlength{\fboxsep}{1pt}\colorbox[RGB]{255,220,170}{\strut\texttt{*\_NUMBERS\_NOT\_SEQ}}}).


\begin{table*}[t]
\centering
\small
\setlength{\tabcolsep}{4pt}
\caption{AssetOpsBench outcome and scale metrics. BASE and SPIN are reported on 261 scenarios; ablations are reported on smaller subsets. Values are mean (std) across runs. Higher is better for Acc; lower is better for Not and Error.}
\label{tab:overall_outcome_scale}
\begin{tabular}{l
  S[table-format=3.0]
  S[table-format=4.0]
  c
  c
  c
}
\toprule
& \multicolumn{2}{c}{Scale} & \multicolumn{3}{c}{Outcome rates (per task)} \\
\cmidrule(lr){2-3}\cmidrule(lr){4-6}
Method & {Runs} & {Tasks} & {Acc} & {Not} & {Error} \\
\midrule
{[BASE]}
& 261 & 1061
& 0.638 (0.289) & 0.244 (0.252) & 0.009 (0.048) \\
{[SPIN]}
& 261 & 623
& \bfseries 0.706 (0.332) & \bfseries 0.195 (0.269) & 0.010 (0.061) \\
{[SPIN\_wo\_sim]}
& 139 & 351
& 0.665 (0.335) & 0.239 (0.294) & 0.008 (0.055) \\
{[SPIN\_wo\_cri]}
& 141 & 405
& 0.619 (0.323) & 0.271 (0.274) & \bfseries 0.007 (0.048) \\
\bottomrule
\end{tabular}
\vspace{-0.1in}
\end{table*}

\begin{table*}[t]
  \centering
  \scriptsize
  \setlength{\tabcolsep}{2pt}
  \renewcommand{\arraystretch}{1.2}
  \caption{AssetOpsBench execution-effort metrics. Values are mean (std) across runs. Lower is better for all metrics, except that token counts also reflect internal prompting overhead.}
  \label{tab:overall_effort}

  \begin{tabular}{lcccccc}
    \toprule
    Method
    & \makecell{AvgTasks\\ /Run}
    & \makecell{ToolCalls\\ /Run}
    & \makecell{ApiCalls\\ /Run}
    & \makecell{TokSent\\ /Run}
    & \makecell{TokRecv\\ /Run}
    & \makecell{Elapsed(s)\\ /Run} \\
    \midrule

    {[BASE]}
    & 4.07 (2.178)
    & 11.81 (7.282)
    & 34.05 (18.655)
    & 111530.3 (109009.698)
    & \bfseries 2590.4 (1609.883)
    & 198.44 (155.552) \\

    {[SPIN]}
    & \bfseries 2.39 (1.153)
    & \bfseries 6.82 (3.990)
    & \bfseries 19.97 (11.175)
    & 117592.7 (92820.875)
    & 5542.3 (6527.553)
    & 143.53 (91.570) \\

    {[SPIN\_wo\_sim]}
    & 2.53 (1.119)
    & 6.94 (4.150)
    & 21.37 (11.856)
    & \bfseries 106564.0 (124601.952)
    & 4033.7 (3135.429)
    & \bfseries 138.41 (77.483) \\

    {[SPIN\_wo\_cri]}
    & 2.87 (1.088)
    & 7.73 (4.249)
    & 23.99 (11.221)
    & 154002.0 (109280.496)
    & 7751.0 (7400.157)
    & 193.54 (103.974) \\
    \bottomrule
  \end{tabular}
\end{table*}

Tables~\ref{tab:overall_outcome_scale} and~\ref{tab:overall_effort} summarize the end-to-end outcome, scale, and execution-effort metrics on AssetOpsBench. Table~\ref{tab:overall_outcome_scale} shows that, on the same 261 scenarios, \texttt{[SPIN]} reduces the total number of executed tasks from 1061 to 623 while increasing task-level \emph{Accomplished} from 0.638 to 0.706. At the same time, \emph{Not} decreases from 0.244 to 0.195, while \emph{Error} remains low and comparable (0.009 for \texttt{[BASE]} vs.\ 0.010 for \texttt{[SPIN]}). These results indicate that the reduction in execution is not merely due to premature stopping or degenerate truncation; rather, SPIN executes fewer tasks while preserving and slightly improving task-level outcome quality.

Table~\ref{tab:overall_effort} shows that this quality improvement is accompanied by a substantial reduction in downstream execution burden. Relative to \texttt{[BASE]}, \texttt{[SPIN]} reduces average tasks per run from 4.07 to 2.39, tool calls per run from 11.81 to 6.82, API calls per run from 34.05 to 19.97, and elapsed runtime from 198.44s to 143.53s. The only major metric that increases is tokens sent per run, from 111{,}530.3 to 117{,}592.7, together with tokens received per run from 2590.4 to 5542.3. This pattern is consistent with the design of SPIN: the method spends additional internal prompting budget on simulator and critic calls in order to avoid more expensive downstream execution. We therefore interpret these results not as a uniform reduction of all computational cost, but as a favorable shift from external execution effort toward internal deliberation.

The ablations help clarify which components drive these improvements. On the smaller ablation subsets, \texttt{[SPIN\_wo\_sim]} remains relatively close to full \texttt{[SPIN]} in task-level outcome rates, with \emph{Accomplished} of 0.665 versus 0.706, and it yields the lowest token-sent value among the SPIN variants at 106{,}564.0 per run. By contrast, weakening critic-based stopping in \texttt{[SPIN\_wo\_cri]} leads to longer and more expensive executions: it processes 405 tasks in total, averages 2.87 tasks per run, makes 23.99 API calls per run, and sends 154{,}002.0 tokens per run. Relative to full \texttt{[SPIN]}, these values suggest that the critic contributes directly to limiting execution length once sufficient evidence has been accumulated.

At the same time, we interpret the ablations cautiously because they are evaluated on smaller subsets than the full 261-scenario comparison. In particular, while Table~\ref{tab:overall_effort} suggests that critic weakening most clearly increases execution length, the matched-scenario failure analysis later shows that removing the simulator more strongly degrades robustness-related failure patterns, especially repetition and other specification/state failures. Taken together, these results suggest a division of labor between the two components: the critic is more directly responsible for stopping control and execution-length reduction, whereas the simulator contributes more strongly to the quality of intermediate state estimation that supports robust execution behavior.

\begin{figure*}[t]
  \centering
  \includegraphics[width=\textwidth]{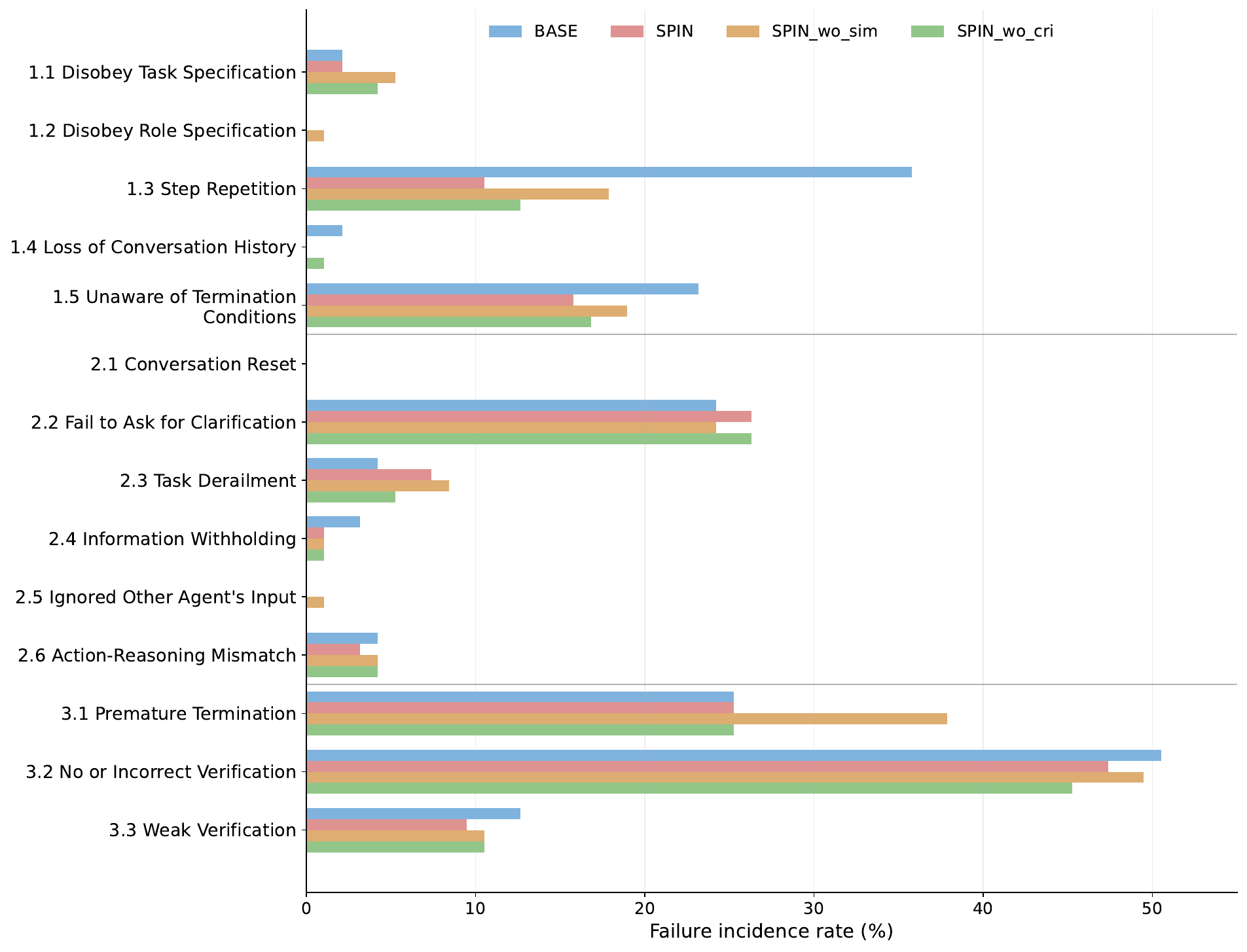}
  \caption{Fine grained failure incidence rates on the exact common 95 scenario intersection across \texttt{[BASE]}, \texttt{[SPIN]}, \texttt{[SPIN\_wo\_sim]}, and \texttt{[SPIN\_wo\_cri]}. Lower is better.}
  \label{fig:fma_full_rate}
\end{figure*}

\subsection{Failure Mode Analysis on AssetOpsBench}
\label{sec:fma_assetopsbench}

Figure~\ref{fig:fma_full_rate} reports fine-grained failure incidence rates on the exact common 95-scenario intersection across \texttt{[BASE]}, \texttt{[SPIN]}, \texttt{[SPIN\_wo\_sim]}, and \texttt{[SPIN\_wo\_cri]}. This analysis complements the aggregate outcome and effort metrics by showing not only whether SPIN improves execution, but also which types of failures are most affected by each component.

Relative to \texttt{[BASE]}, full \texttt{[SPIN]} most clearly reduces specification/state-related failures, especially \texttt{1.3 Step Repetition}, and also lowers \texttt{1.5 Unaware of Termination Conditions}. This pattern is consistent with the intended role of prefix-based execution control: SPIN suppresses repeated or unnecessarily prolonged execution and improves stopping behavior once the current prefix is already sufficient. In other words, the main robustness gain of SPIN appears in reducing redundant continuation rather than in repairing every downstream failure category equally.

At the same time, the figure also shows an important limitation. Failure types such as \texttt{2.2 Fail to Ask for Clarification} and \texttt{3.2 No or Incorrect Verification} are not clearly improved by full \texttt{[SPIN]}. One plausible interpretation is that SPIN is explicitly optimized to remove steps that look unnecessary for answering the query under its current internal estimate of sufficiency. As a result, steps that would have acted as extra safeguards against missing information or weak verification can also be pruned away as redundant. From this perspective, SPIN becomes more efficient and less repetitive, but does not necessarily become more conservative with respect to clarification or verification failures. Thus, the FMA suggests that SPIN improves execution efficiency and progression control more than it improves robustness to uncertainty-handling or verification-related errors.

The ablations further clarify component roles. Removing the simulator causes the failure profile to regress toward \texttt{[BASE]}, especially for \texttt{1.3 Step Repetition} and \texttt{3.1 Premature Termination}, whereas removing the critic leaves a profile that remains comparatively closer to full \texttt{[SPIN]}. This suggests that the simulator is the more critical component for robust intermediate state estimation and stable execution behavior on the matched scenario set, while the critic contributes more directly to execution-length control, as also suggested by the end-to-end effort metrics.

\subsection{MCP Bench results}
\label{sec:mcp_results}

Table~\ref{tab:mcpbench_metrics} reports results on 18 two-server grounding tasks from MCP Bench. We evaluate Baseline and SPIN-wrapped runs for GPT-OSS1 (ID~20) and Llama~4~Maverick (ID~16), using the model mapping in Table~\ref{tab:llm_id_to_model}. We include MCP Bench for a different reason than AssetOpsBench: in this setting, we do not use any trajectory database or external retrieval memory. As a result, MCP Bench tests whether SPIN, and in particular its simulator-centered prefix evaluation, can remain effective when the relevant task knowledge must come primarily from the base LLM's prior knowledge rather than from benchmark-specific stored experience.

Under this no-external-memory setting, SPIN still improves most of the quality-oriented metrics for both models. For GPT-OSS1, Task Completion increases from 2.39 to 3.80, Tool Selection from 2.99 to 3.56, Planning Effectiveness from 2.04 to 2.74, Grounding from 2.44 to 3.21, Dependency Awareness from 2.39 to 3.49, and Parallelism \& Efficiency from 1.69 to 2.00. For Llama~4~Maverick, Task Completion increases from 2.33 to 2.86, Tool Selection from 2.73 to 3.43, Planning Effectiveness from 1.99 to 2.22, Grounding from 2.54 to 3.19, Dependency Awareness from 2.28 to 2.63, and Parallelism \& Efficiency from 1.70 to 1.81. These gains are important because they show that the benefits of SPIN are not limited to settings where the simulator can draw on an external trajectory store. Instead, the same wrapper continues to improve planning and grounding behavior even when its intermediate judgments must be supported primarily by the LLM's internal prior knowledge.

The efficiency effects, however, are model-dependent. For GPT-OSS1, Avg Rounds remains unchanged at 7.06, and Prompt Tokens increase slightly from 55{,}464.78 to 56{,}890.44. By contrast, for Llama~4~Maverick, SPIN reduces Avg Rounds from 17.06 to 14.00 and Prompt Tokens from 212{,}118.22 to 125{,}243.06. We interpret this difference as evidence that SPIN provides a useful control mechanism for both models, but that the magnitude of the efficiency benefit depends on the base model's native planning and execution behavior. In other words, SPIN improves the quality of plan progression in both cases, while the extent to which this translates into shorter interaction traces or lower token cost is model-specific.

One exception is Tool Call Success for GPT-OSS1, which decreases slightly from 0.97 to 0.92. We therefore do not claim that SPIN uniformly improves every low-level execution metric. Rather, the overall pattern suggests that SPIN consistently improves higher-level planning and grounding quality, while some lower-level tool-execution behaviors may remain sensitive to the underlying model. Taken together, the MCP-Bench results support a stronger claim than simple benchmark transfer: even without any database-backed external memory, SPIN remains effective when the task can be supported by the LLM's prior knowledge alone.

\begin{table*}[t]
\centering
\small
\setlength{\tabcolsep}{4pt}
\caption{MCP-Bench results on 18 two-server grounding tasks. Rows are model variants with internal LLM IDs: GPT-OSS1 (ID~20) and Llama~4~Maverick (ID~16) (Table~\ref{tab:llm_id_to_model}); each is reported for Baseline and SPIN-wrapped runs. Columns report core quality metrics (higher is better) and efficiency/cost proxies (lower is better). Best values are in bold (ties allowed).}
\vspace{-0.1in}
\label{tab:mcpbench_metrics}
\resizebox{\textwidth}{!}{%
\begin{tabular}{lccccccccc}
\toprule
& \multicolumn{7}{c}{Quality}
& \multicolumn{2}{c}{Efficiency / Cost} \\
\cmidrule(lr){2-8}\cmidrule(lr){9-10}
Model
& \makecell{Task\\Completion}
& \makecell{Tool\\Selection}
& \makecell{Planning\\Effect.}
& Grounding
& \makecell{Tool Call\\Success}
& \makecell{Dependency\\Awareness}
& \makecell{Parallelism\\\& Eff.}
& \makecell{Avg\\Rounds}
& \makecell{Prompt\\Tokens} \\
\midrule
GPT-OSS1 (SPIN)
& \bfseries 3.80 & \bfseries 3.56 & \bfseries 2.74 & \bfseries 3.21
& 0.92 & \bfseries 3.49 & \bfseries 2.00 & \bfseries 7.06 & 56890.44 \\
GPT-OSS1 (Baseline)
& 2.39 & 2.99 & 2.04 & 2.44
& \bfseries 0.97 & 2.39 & 1.69 & \bfseries 7.06 & \bfseries 55464.78 \\
Llama 4 Maverick (SPIN)
& 2.86 & 3.43 & 2.22 & 3.19
& 0.83 & 2.63 & 1.81 & 14.00 & 125243.06 \\
Llama 4 Maverick (Baseline)
& 2.33 & 2.73 & 1.99 & 2.54
& 0.79 & 2.28 & 1.70 & 17.06 & 212118.22 \\
\bottomrule
\end{tabular}
}
\end{table*}

\section{Limitations}
Our evaluation is limited to AssetOpsBench and its fixed tool/agent environment, so the observed quality and efficiency trade-offs needs testing to other benchmarks, or tool APIs with different latency and failure characteristics.

In addition, the simulator depends on the coverage, freshness, and representativeness of logged trajectories; sparse or shifting environments can yield inaccurate simulated outcomes that may propagate into suboptimal intermediate decisions. Early stopping further relies on an LLM-based critic, which can be miscalibrated (false positives halting too early; false negatives reducing efficiency) and is sensitive to rubric design, few-shot selection, and model choice, with no formal guarantees on stopping correctness. Finally, SPIN introduces additional components (repair prompting, database retrieval, simulator, critic) that increase engineering complexity and per-step inference overhead.

\section{Conclusion}
\label{sec:conclusion}

We presented SPIN, a planning wrapper for tool-using LLM agents that combines executable DAG validation with prefix-based execution control. Across our experiments, SPIN improved planner outputs as machine-consumable interfaces and reduced downstream execution burden while maintaining or improving task quality. On AssetOpsBench, this appeared as both better structural executability and better end-to-end execution efficiency, including fewer executed tasks, tool calls, API calls, and shorter runtimes, while task-level \emph{Accomplished} also improved. The ablation and failure-mode analyses further suggest that these gains arise primarily from better progression and stopping control: SPIN most clearly reduces redundant continuation and repetition-related failures, while the simulator and critic play different roles in robust state estimation and execution-length control.

At the same time, our results also identify an important boundary of the current approach. Clarification- and verification-related failures are not substantially improved, suggesting that SPIN is more effective at pruning unnecessary continuation than at adding robustness-oriented safeguards under uncertainty. A natural next step is therefore to extend SPIN toward conditional planning with explicit clarification and fallback branches. More broadly, an interesting future direction is to use the simulator as a world model that exposes recurring failure patterns and enables the LLM to learn, through few-shot examples, when a task appears risky before execution. This could allow the system not only to stop earlier, but also to recognize in advance when additional clarification, verification, or alternative branches are needed.

\bibliographystyle{plainnat}
\bibliography{custom}


\appendix

\appendix
\section{Reproducibility Details}
\label{app:repro}

\subsection{Experiment Regeneration}
\label{sec:experiment_regeneration}

\paragraph{Deterministic table recomputation from saved run artifacts.}
All tables reported in this paper are computed \emph{solely} from the local artifacts stored under
AssetOpsBench/benchmark/cods\_track1/track1\_result \citep{codabench_assetopsbench_challenge}, without re-running the benchmark execution or invoking any external APIs.
In particular, the table generation scripts read (i) step level execution traces under
\texttt{track1\_result/trajectory} and (ii) per experiment metadata and summaries under
\texttt{track1\_result/exp}.
Given a fixed \texttt{track1\_result} directory, this post processing stage is deterministic and therefore reproduces the same aggregate numbers exactly.
For example, Table~2 can be regenerated as follows:
\begin{Verbatim}[fontsize=\scriptsize,breaklines=true,breakanywhere=true]
AssetOpsBench % python3 make_table_2.py \
  --trajectory_root "./benchmark/cods_track1/track1_result/trajectory" \
  --exp_root        "./benchmark/cods_track1/track1_result/exp" \
  --model "Model_16" \
  --tags  "BASE,SPIN,SPIN_wo_sim,SPIN_wo_cri" \
  --out_dir "./benchmark/cods_track1/track1_result/tables3" \
  --debug
\end{Verbatim}

\paragraph{end to end trajectory regeneration via Docker Compose.}
In addition to the deterministic post processing described above, we provide an end to end replay path that regenerates \texttt{trajectory} (and per-run \texttt{exp} summaries) by re-running the benchmark inside a Docker Compose environment. This replay depends on external model/API calls; therefore, we do not guarantee bitwise identical outputs or exact metric equality across runs. Instead, we expect the aggregate statistics (e.g., success rate and error type distributions) to be broadly consistent when the same scenario subset and configuration are used.

\smallskip
\noindent\textbf{Configuration.}
Users must create \texttt{benchmark/cods\_track1/.env.local} to provide the required API credentials and endpoints. Secrets must remain local and must not be committed.

\smallskip
\noindent\textbf{Run (Compose).}
\begin{lstlisting}[language=bash,basicstyle=\ttfamily\footnotesize,breaklines=true]
# From the repository root (AssetOpsBench/)
docker compose -f benchmark/cods_track1/docker-compose.yml up -d --build

# (Optional) If you run the benchmark command explicitly in the container:
# docker compose -f benchmark/cods_track1/docker-compose.yml exec -T assetopsbench \
#   python /home/run_track_1.py --utterance_ids "<comma-separated IDs>" --save_data "True"
\end{lstlisting}

\smallskip
\noindent\textbf{Outputs.}
The end to end run writes local artifacts under
\texttt{benchmark/cods\_track1/track1\_result/trajectory} and
\texttt{benchmark/cods\_track1/track1\_result/exp}. All tables in this paper are computed solely from these saved artifacts without re-invoking external APIs.

\paragraph{Failure mode analysis replay via TrajFM.}
In addition to the benchmark trajectory regeneration above, we provide a replay path for the failure mode analysis (FMA) used in the appendix discussion. The FMA pipeline runs the TrajFM extractor separately on saved trajectory directories for \texttt{[BASE]}, \texttt{[SPIN]}, \texttt{[SPIN\_wo\_sim]}, and \texttt{[SPIN\_wo\_cri2]}, and writes the resulting per-method outputs under \texttt{track1\_result/trajfm\_outputs/}. This stage depends on external model/API calls and therefore is not guaranteed to be bitwise identical across replays, although the resulting failure distributions are expected to be broadly consistent when the same trajectory inputs and evaluator configuration are used. For consistency, we refer to this variant as \texttt{[SPIN\_wo\_cri]} in the paper, while the corresponding artifact directory is named \texttt{spin\_wo\_cri2/work/} in the released outputs.

\smallskip
\noindent\textbf{Run (TrajFM).}
The extraction step is executed inside the benchmark container with:
\begin{lstlisting}[language=bash,basicstyle=\ttfamily\footnotesize,breaklines=true]
# From the repository root (AssetOpsBench/)
docker compose -f benchmark/cods_track1/docker-compose.yml exec -T assetopsbench \
  bash /home/entrypoint_failure_modes_analysis.sh
\end{lstlisting}

This entrypoint activates the benchmark environment, reads saved trajectories from
\texttt{/home/track1\_result/trajectory/}, and runs
\texttt{TrajFM/failure\_mode\_extractor.py} separately for each method directory.

\smallskip
\noindent\textbf{Outputs.}
The FMA extraction writes method specific outputs under:
\begin{Verbatim}[fontsize=\scriptsize,breaklines=true,breakanywhere=true]
benchmark/cods_track1/track1_result/trajfm_outputs/
  base/work/
  spin/work/
  spin_wo_sim/work/
  spin_wo_cri2/work/
\end{Verbatim}
These directories contain the raw TrajFM outputs used for the matched scenario analysis in Appendix~\ref{app:fma_details}.

\smallskip
\noindent\textbf{Figure generation.}
After the FMA counts and rates are computed, the plotting script
\texttt{make\_fma\_figure.py} generates the rate figures used in the paper:
\begin{lstlisting}[language=bash,basicstyle=\ttfamily\footnotesize,breaklines=true]
# From the repository root (AssetOpsBench/)
python make_fma_figure.py
\end{lstlisting}
This script writes:
\begin{Verbatim}[fontsize=\scriptsize,breaklines=true,breakanywhere=true]
benchmark/cods_track1/track1_result/trajfm_outputs/fma_full_rate.pdf
benchmark/cods_track1/track1_result/trajfm_outputs/fma_category_rate.pdf
\end{Verbatim}
The current plotting script renders figures from precomputed aggregated rates; it does not itself perform the raw trajectory labeling or aggregation.

\smallskip
\noindent\textbf{External API dependency note.}
Our model routing / numeric model identifiers for external calls follow the implementation in
ReActXen's \texttt{model\_inference.py}\footnote{\url{https://github.com/IBM/ReActXen/blob/main/src/reactxen/utils/model_inference.py}}.
If the provider changes the identifier mapping or discontinues service, exact replay may become impossible, and only partial reproducibility from saved artifacts is guaranteed.

\subsection{Submission Package Structure (Software Archive / Data Archive)}
\label{app:repro_package}

\paragraph{Overview.}
We provide two archives for reproducibility:
(i) a software archive (\texttt{AssetOpsBench.zip}) containing a code snapshot of
AssetOpsBench (https://anonymous.4open.science/r/AssetOpsBench-2C61) \citep{codabench_assetopsbench_challenge}, and
(ii) a data archive (\texttt{pgdata.tgz}) containing a populated PostgreSQL
data directory (\texttt{PGDATA}).
The CODS Track~1 Docker Compose stack connects services on a default Compose network,
so containers can reach each other by service name (e.g., \texttt{postgres}, \texttt{couchdb}). 

\paragraph{A. Archive metadata (minimum).}
\begin{itemize}
  \item \textbf{Software archive filename (zip/tgz):} \texttt{AssetOpsBench.zip}
  \item \textbf{Data archive filename (zip/tgz):} \texttt{pgdata.tgz}
\end{itemize}

\paragraph{B. Top-level contents (minimum).}
\noindent\textbf{Software archive: top-level directory listing (depth $\le 2$).}
\begin{verbatim}
AssetOpsBench/
  benchmark/
    cods_track1/
  infra/
    postgres/
  src/
    agent_hive/
  readme.txt
  (other files/directories omitted)
\end{verbatim}

\noindent\textbf{Data archive: top-level directory listing (depth $\le 2$).}
\begin{verbatim}
pgdata.tgz (archive root)/
  data/
    PG_VERSION
    base/
    global/
    pg_wal/
    pg_notify/
    (other Postgres subdirectories omitted)
\end{verbatim}

\paragraph{Key implementation location.}
The primary workflow implementation used in our Track~1 experiments is under
\texttt{src/agent\_hive/workflows/} (e.g., planning workflows, critic/simulator modules,
and Track~1 workflow variants). For reference, this directory contains modules such as
\texttt{track1\_planning\_baseline.py}, \texttt{track1\_planning\_spin.py},
\texttt{track1\_planning\_validation.py}, \texttt{critic\_agent.py},
\texttt{simulator\_agent.py}, and \texttt{validate\_plan\_text.py}.

\paragraph{Configuration switch via \texttt{docker-compose.yml}.}
In \texttt{benchmark/cods\_track1/docker-compose.yml}, the execution configuration can be
switched by changing the suffixes of three referenced files:
\texttt{<FILE\_A>}, \texttt{<FILE\_B>}, and \texttt{<FILE\_C>}.
In our setup, these references select which Track~1 workflow variant (e.g., baseline vs.\ SPIN variants)
is executed. Reviewers can reproduce different configurations by editing these three filenames in the
Compose file and restarting the services.

\paragraph{Restoration and rerun instructions (readme.txt).}
The software archive includes a repository-level instruction file at
\texttt{AssetOpsBench/readme.txt}. Following \texttt{readme.txt}, reviewers can:
(i) restore the Postgres state by extracting \texttt{pgdata.tgz} into \texttt{infra/postgres/},
(ii) recreate the required local environment file \texttt{benchmark/cods\_track1/.env.local}, and
(iii) rerun the Compose stack and regenerate the tables (Table~1 via \texttt{make\_dag\_table.py} and
Table~2 via \texttt{make\_table\_2.py}).

\paragraph{Database restoration note.}
To restore the database directory to \texttt{infra/postgres/data}, place \texttt{pgdata.tgz} at the
repository root and extract using \texttt{-C} (change directory before extraction): 
\begin{verbatim}
tar -xzf pgdata.tgz -C infra/postgres data
\end{verbatim}

\paragraph{Environment file note.}
The Compose configuration expects \texttt{benchmark/cods\_track1/.env.local} to exist before
\texttt{docker compose} is executed; missing env files can trigger a startup error unless the Compose
configuration explicitly marks them optional.

\subsection{Fixed Experimental Settings (Configuration Only)}
\label{app:fixed_settings_config}

\paragraph{Evaluation scenario sets (fixed).}
We use different fixed evaluation sets for different parts of the paper.
For the offline DAG validation study in Table~\ref{tab:dag_success_validate_plan_text}, we evaluate 141 plan files per planner configuration.
For the end to end AssetOpsBench results in Table~\ref{tab:overall_outcome_scale} and Table~\ref{tab:overall_effort}, we evaluate 261 scenarios.
For the MCP Bench results in Table~\ref{tab:mcpbench_metrics}, we evaluate 18 two server grounding tasks.
Within each table, all reported aggregates are computed over the corresponding fixed scenario set.

\paragraph{Outcome sensitive knobs (fixed).}
The reported aggregates depend on the following configuration choices:
\begin{itemize}
  \item \textbf{SPIN iteration / validation budget ($T$).}
  The number of SPIN iterations is controlled in
  \texttt{AssetOpsBench/src/agent\_hive/workflows/\\track1\_planning\_\{...\}.py}
  (denoted as \texttt{T=} in the code). Changing $T$ changes both cost and outcomes.

  \item \textbf{Simulator retrieval filter.}
  The Simulator retrieves task summaries restricted to \texttt{Accomplished} entries,
  that is, only completed historical tasks are used as retrieval candidates.
  Changing this filter, for example by including \texttt{Not accomplished} cases,
  may change Simulator outputs.

  \item \textbf{Early stopping rule (Critic $\rightarrow$ stop).}
  We enable early stopping when the Critic returns
  \texttt{can\_answer\_now=True} for tasks that the Simulator predicts as
  \texttt{Accomplished} or \texttt{Partially\_Accomplished}.
  Restricting the condition to only \texttt{Accomplished} is an alternative design choice
  and can change outcomes.

  \item \textbf{Data persistence and feedback effects (\texttt{--save\_data}).}
  When \texttt{--save\_data} is enabled, generated plans and trajectories are written to
  the database and become queryable by downstream components.
  This can affect later Simulator retrieval behavior, and thus end to end results,
  by changing the database state seen by retrieval.
\end{itemize}

\paragraph{External API conditions (model identification and logging).}
When reproducing results that require external LLM APIs, we identify the exact
provider and model via our internal \texttt{LLM\_ID}.
Table~\ref{tab:llm_id_to_model} maps each \texttt{LLM\_ID} to the corresponding
model string used in the implementation, together with provider and deployment details.
For each external API call, we log (i) timestamp, (ii) \texttt{LLM\_ID} and provider/model identifier,
(iii) inference parameters such as temperature, top-$p$, and max\_tokens, and
(iv) retry and timeout metadata such as retry count and timeout type.
These records enable post hoc verification of what could vary across runs,
including provider model updates, stochastic decoding, and rate limit induced retries or timeouts.

\subsection{Failure mode analysis on matched scenarios}
\paragraph{Scope and comparison setting.}
We conduct failure mode analysis on AssetOpsBench trajectories from \texttt{[BASE]}, \texttt{[SPIN]}, \texttt{[SPIN\_wo\_sim]}, and \texttt{[SPIN\_wo\_cri]}. To ensure fair comparison, we restrict the analysis to the exact common utterance ID intersection across all four methods. After collapsing to one trajectory per utterance ID, this yields 95 shared scenarios.

\paragraph{Model and evaluation prompt.}
Failure labels were assigned by an LLM based evaluator using internal \texttt{LLM\_ID}~12 (Table~\ref{tab:llm_id_to_model}). For each trajectory, the evaluator receives the user question, the executed trajectory, and the final answer, and returns a structured multi label judgement over the predefined failure taxonomy described below.

\paragraph{Failure taxonomy.}
We use a three category failure taxonomy covering specification and state related failures (1.x), interaction failures (2.x), and verification and completion failures (3.x). Table~\ref{tab:fma_taxonomy} defines all failure modes used in the analysis.

\begin{table*}[t]
\centering
\small
\setlength{\tabcolsep}{4pt}
\renewcommand{\arraystretch}{1.12}
\caption{Failure taxonomy used in the matched scenario failure mode analysis.}
\label{tab:fma_taxonomy}
\begin{tabularx}{\textwidth}{@{} J{1.1cm} J{3.2cm} Q @{}}
\toprule
\textbf{ID} & \textbf{Category} & \textbf{Definition} \\
\midrule
1.1 & Specification and state & \textbf{Disobey Task Specification}: the trajectory does not follow the requested task or violates explicit task constraints. \\
1.2 & Specification and state & \textbf{Disobey Role Specification}: the agent acts outside its assigned role or uses behavior inconsistent with the role definition. \\
1.3 & Specification and state & \textbf{Step Repetition}: the trajectory repeats the same or nearly identical step without making meaningful progress. \\
1.4 & Specification and state & \textbf{Loss of Conversation History}: the agent ignores or forgets relevant context established earlier in the interaction. \\
1.5 & Specification and state & \textbf{Unaware of Termination Conditions}: the agent fails to recognize when the task should stop or when enough information has already been obtained. \\
\midrule
2.1 & Interaction & \textbf{Conversation Reset}: the trajectory restarts the interaction or behaves as if prior turns were unavailable. \\
2.2 & Interaction & \textbf{Fail to Ask for Clarification}: the agent proceeds without requesting clarification even though the available information is insufficient or ambiguous. \\
2.3 & Interaction & \textbf{Task Derailment}: the interaction drifts away from the intended objective and focuses on irrelevant or secondary subproblems. \\
2.4 & Interaction & \textbf{Information Withholding}: relevant information that was obtained is not surfaced in the response or not used when it should be. \\
2.5 & Interaction & \textbf{Ignored Other Agent's Input}: outputs or observations from previous agents or earlier steps are not incorporated when they are relevant to the next action. \\
2.6 & Interaction & \textbf{Action--Reasoning Mismatch}: the stated reasoning and the executed action are inconsistent with each other. \\
\midrule
3.1 & Verification and completion & \textbf{Premature Termination}: the trajectory ends before enough evidence has been collected to support the answer. \\
3.2 & Verification and completion & \textbf{No or Incorrect Verification}: the agent does not verify an important claim, or performs a verification step incorrectly. \\
3.3 & Verification and completion & \textbf{Weak Verification}: verification is attempted, but it is incomplete, shallow, or insufficiently connected to the final conclusion. \\
\bottomrule
\end{tabularx}
\end{table*}

\section{Failure Mode Analysis Details}
\label{app:fma_details}

\subsection{Failure mode analysis on matched scenarios}
\label{app:fma_matched}

\paragraph{Scope and comparison setting.}
We conduct failure mode analysis on AssetOpsBench trajectories from \texttt{[BASE]}, \texttt{[SPIN]}, \texttt{[SPIN\_wo\_sim]}, and \texttt{[SPIN\_wo\_cri]}. To ensure fair comparison, we restrict the analysis to the exact common utterance ID intersection across all four methods. After collapsing to one trajectory per utterance ID, this yields 95 shared scenarios.

\paragraph{Model and evaluation prompt.}
Failure labels are assigned by an LLM based evaluator using a fixed prompt and a predefined failure taxonomy. We use the same evaluation setup across all four methods so that differences in reported failure rates reflect differences in the trajectories rather than differences in the labeling procedure. We used model 12 as the evaluator (Table~\ref{tab:llm_id_to_model}). 

\paragraph{Aggregation and reporting protocol.}
For each utterance ID, we aggregate failure labels across repeated annotations by OR aggregation, so that a failure mode is counted as present if it appears at least once for that scenario. We report both fine grained failure rates over the full taxonomy and category level aggregates over 1.x, 2.x, and 3.x groups. We also report the mean number of failure modes per scenario on the same matched scenario set.

\paragraph{Results on the matched scenario set.}
Figure~\ref{fig:fma_full_rate} report failure incidence rates on the exact common 95 scenario intersection across \texttt{[BASE]}, \texttt{[SPIN]}, \texttt{[SPIN\_wo\_sim]}, and \texttt{[SPIN\_wo\_cri]}. Under this strict matched comparison, full \texttt{[SPIN]} has the lowest mean number of failure modes per scenario at 1.48, compared with 1.87 for \texttt{[BASE]}, 1.80 for \texttt{[SPIN\_wo\_sim]}, and 1.53 for \texttt{[SPIN\_wo\_cri]}.

The clearest differences appear in 1.x specification and state related failures. In category totals, 1.x decreases from 60 for \texttt{[BASE]} to 27 for full \texttt{[SPIN]}, while \texttt{[SPIN\_wo\_sim]} and \texttt{[SPIN\_wo\_cri]} have 41 and 33, respectively. Within this category, the largest single reduction is \texttt{1.3 Step Repetition}, which drops from 35.79\% in \texttt{[BASE]} to 10.53\% in full \texttt{[SPIN]}; the corresponding rates are 17.89\% for \texttt{[SPIN\_wo\_sim]} and 12.63\% for \texttt{[SPIN\_wo\_cri]}. \texttt{1.5 Unaware of Termination Conditions} also decreases from 23.16\% in \texttt{[BASE]} to 15.79\% in full \texttt{[SPIN]}.

For 3.x verification and completion failures, the category totals are 84 for \texttt{[BASE]}, 78 for full \texttt{[SPIN]}, 93 for \texttt{[SPIN\_wo\_sim]}, and 77 for \texttt{[SPIN\_wo\_cri]}. At the fine grained level, \texttt{3.2 No or Incorrect Verification} remains high for all methods, but is lower for full \texttt{[SPIN]} than for \texttt{[BASE]} (47.37\% vs.\ 50.53\%). \texttt{3.3 Weak Verification} is also lower for full \texttt{[SPIN]} than for \texttt{[BASE]} (9.47\% vs.\ 12.63\%). In contrast, \texttt{3.1 Premature Termination} is identical for \texttt{[BASE]}, full \texttt{[SPIN]}, and \texttt{[SPIN\_wo\_cri]} at 25.26\%, while \texttt{[SPIN\_wo\_sim]} is higher at 37.89\%.

The 2.x interaction failures vary less across methods. The category totals are 34 for \texttt{[BASE]}, 36 for full \texttt{[SPIN]}, 37 for \texttt{[SPIN\_wo\_sim]}, and 35 for \texttt{[SPIN\_wo\_cri]}. In particular, \texttt{2.2 Fail to Ask for Clarification} is 24.21\% for \texttt{[BASE]}, 26.32\% for full \texttt{[SPIN]}, 24.21\% for \texttt{[SPIN\_wo\_sim]}, and 26.32\% for \texttt{[SPIN\_wo\_cri]}.


\begin{table*}[t]
\centering
\scriptsize
\setlength{\tabcolsep}{3pt}
\renewcommand{\arraystretch}{1.05}

\caption{Mapping from our internal LLM IDs to the model strings used in ReActXen's \texttt{model\_inference.py}, together with provider and deployment notes. Unless explicitly stated, quantization and low level serving optimizations are provider managed and not disclosed. Decoding parameters follow our implementation defaults when no overrides are provided: temperature $= 0.0$ and max tokens $= 8192$ for text generation.}
\label{tab:llm_id_to_model}

\begin{tabularx}{\textwidth}{@{}C{0.7cm} L{4.6cm} L{1.2cm} L{2.6cm} C{1.5cm} X@{}}
\toprule
\textbf{ID} & \textbf{Model string in ReActXen} & \textbf{Provider} & \textbf{Size (total / active)} & \textbf{Decoding (temp / max tok)} & \textbf{Quantization / Hosting} \\
\midrule
12 & \path{meta-llama/llama-3-3-70b-instruct} & Meta & 70B (dense) & 0.0 / 8192 & Provider default; self hostable open weight \\
16 & \path{meta-llama/llama-4-maverick-17b-128e-instruct-fp8} & Meta & 400B total / 17B active & 0.0 / 8192 & Provider default; self hostable (license gated) \\
20 & \path{openai/gpt-oss-120b} & OpenAI & 117B total / 5.1B active & 0.0 / 8192 & Provider default; self hostable open weight \\
22 & \path{mistralai/mistral-medium-2505} & Mistral AI & (see provider docs) & 0.0 / 8192 & Provider default; hosted API \\
23 & \path{mistralai/mistral-small-3-1-24b-instruct-2503} & Mistral AI & 24B (dense) & 0.0 / 8192 & Provider default; open weight + hosted options \\
38 & \path{ibm/granite-4-h-small} & IBM & 32B total / 9B active & 0.0 / 8192 & Provider default; self hostable open weight \\
\bottomrule
\end{tabularx}
\end{table*}

\section{Working Examples}
\label{sec:working_examples}
\subsection{Various DAG examples}
\label{sec:dag_examples}
\begin{figure*}[t!]
  \centering

  \begin{tcolorbox}[
    enhanced,
    colback=white,
    colframe=black!20,
    boxrule=0.5pt,
    arc=1.5mm,
    left=2mm,right=2mm,top=1.5mm,bottom=1.5mm,
    width=\textwidth
  ]
  \small
  \begin{Verbatim}[
    breaklines=true,
    breakanywhere=true,
    breaksymbolleft={},
    breaksymbolright={},
    commandchars=\\\{\}
  ]
Question: After checking existing work orders, anomalies and alerts in May and June 2020 for Chiller 9 with equipment ID CWC04009, should I recommend a new work order or is it too early to decide for July 2020?
Plan:
\HLcyan{#Task1: Retrieve existing work orders for Chiller 9 with equipment ID CWC04009 in May and June 2020.}
\HLcyan{#Agent1: WorkOrder Agent}
\HLcyan{#Dependency1: None}
\HLcyan{#ExpectedOutput1: List of existing work orders for Chiller 9 (CWC04009) in May and June 2020.}

\HLred{#Task2: Check for anomalies and alerts for Chiller 9 (CWC04009) in May and June 2020.}
\HLred{#Agent2: Time Series Analytics and Forecasting}
\HLred{#Dependency2: #S1}
\HLred{#ExpectedOutput2: Analysis of anomalies and alerts for Chiller 9 (CWC04009) in the specified period.}

\HLorange{#Task3: Analyze the results from existing work orders and anomalies/alerts to determine if a new work order is needed for July 2020.}
\HLorange{#Agent3: Failure Mode and Sensor Relevancy Expert for Industrial Asset}
\HLorange{#Dependency3: #S1, #S2}
\HLorange{#ExpectedOutput3: Recommendation on whether to create a new work order for Chiller 9 (CWC04009) in July 2020.}

\HLgreen{#Task4: Create a new work order if recommended by the Failure Mode and Sensor Relevancy Expert.}
\HLgreen{#Agent4: WorkOrder Agent}
\HLgreen{#Dependency4: #S3}
\HLgreen{#ExpectedOutput4: New work order created if necessary.}
  \end{Verbatim}
  \end{tcolorbox}

  \vspace{0.8ex}

  \includegraphics[
    width=\textwidth,
    height=0.38\textheight,
    keepaspectratio
  ]{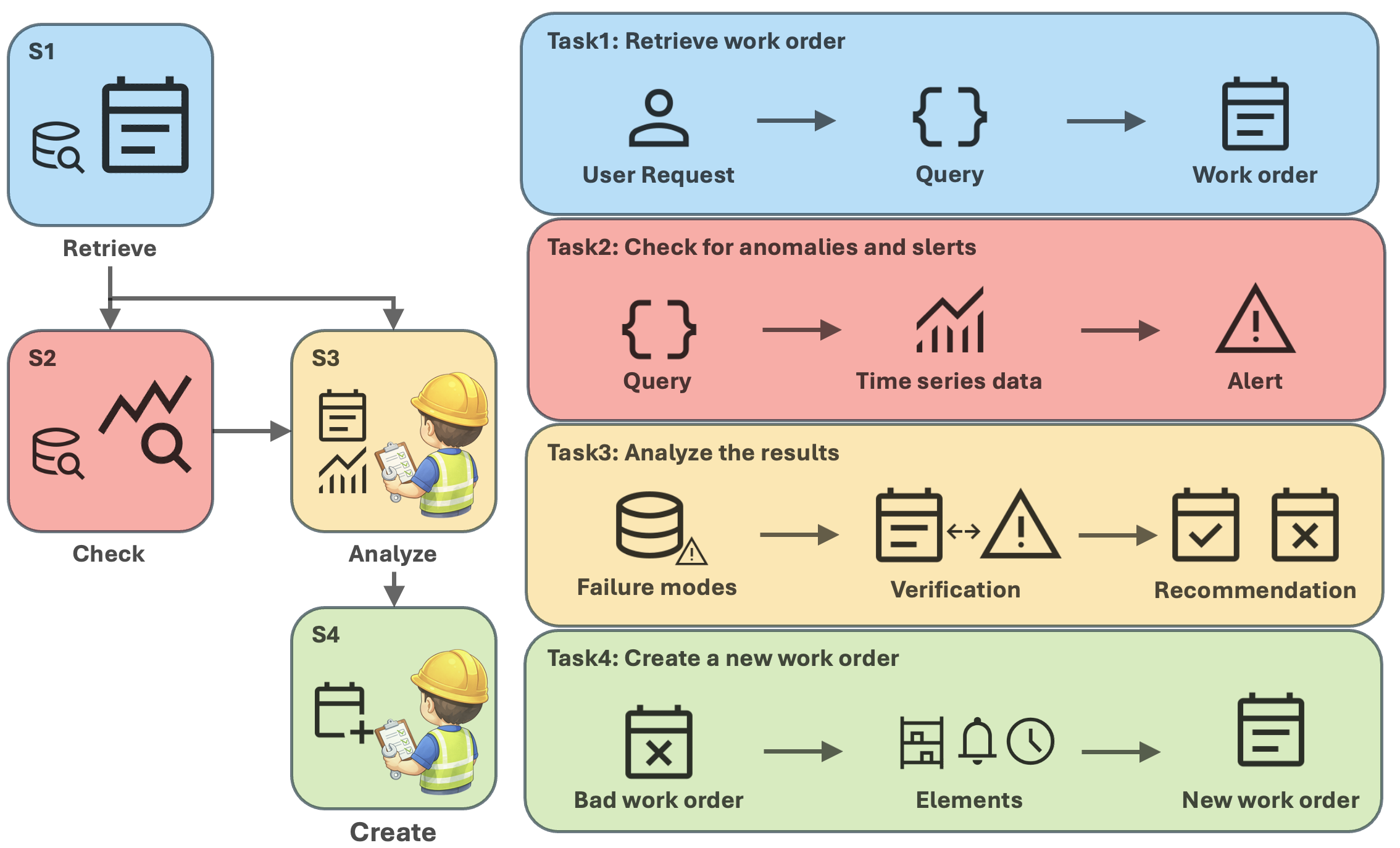}

  \caption{
Visualization of a four-step DAG plan for a maintenance decision on Chiller~9 (equipment ID: CWC04009).
S1--S4 denote the DAG step indices referenced by the plan dependencies (\#S1--\#S4), corresponding to Tasks~1--4, respectively.
The left panel shows the step-level dependency graph (arrows indicate prerequisite relations).
The right panel summarizes each step as a compact workflow (\emph{query} $\rightarrow$ \emph{retrieve/analyze} $\rightarrow$ \emph{output}). Each workflow is executed by its designated agent, which attempts to complete the task using the tools available to it. The color coding matches the highlighted blocks in the plan text.}
  \label{fig:dag_visual}
\end{figure*}

\begin{table*}[t]
  \centering
  \scriptsize

  \caption{Four representative DAG plan examples (left: plan text; right: corresponding dependency structure).}
  \label{tab:dag_examples}

  \noindent
  \begin{minipage}[t]{\DAGLeftW}
    \hrule height 0pt 
    \textbf{DAG plan (text)}
  \end{minipage}\hfill
  \begin{minipage}[t]{\DAGRightW}
    \hrule height 0pt 
    \centering\textbf{Mini DAG (structure)}
  \end{minipage}

  \DAGSep

  \noindent
  \begin{minipage}[t]{\DAGLeftW}
    \hrule height 0pt 
    \begin{tcolorbox}[
      enhanced, breakable,
      colback=white, colframe=black!25,
      boxrule=0.35pt, arc=1.2mm,
      boxsep=0.3mm, left=0.6mm, right=0.6mm, top=0.3mm, bottom=0.3mm
    ]
    \ttfamily\RaggedRight
\begin{DAGPlanVrb}
#Task1: Identify the relevant sensors for monitoring compressor overheating failure in Chiller 6.
#Agent1: Failure Mode and Sensor Relevancy Expert for Industrial Asset
#Dependency1: None
#ExpectedOutput1: List of relevant sensors for monitoring compressor overheating in Chiller 6.

#Task2: Prioritize the identified sensors for monitoring compressor overheating in Chiller 6.
#Agent2: Failure Mode and Sensor Relevancy Expert for Industrial Asset
#Dependency2: #S1
#ExpectedOutput2: The most relevant sensor to prioritize for monitoring compressor overheating in Chiller 6.
\end{DAGPlanVrb}
    \end{tcolorbox}
  \end{minipage}\hfill
  \begin{minipage}[t]{\DAGRightW}
    \hrule height 0pt 
    \centering
    \textbf{Path}\par\vspace{-0.4ex}
    {\tiny
    \begin{tabular}{c}
      \fbox{\strut S1}\\[-0.4ex]
      $\downarrow$\\[-0.4ex]
      \fbox{\strut S2}
    \end{tabular}}
  \end{minipage}

  \DAGSep

  \noindent
  \begin{minipage}[t]{\DAGLeftW}
    \hrule height 0pt
    \begin{tcolorbox}[enhanced, breakable,
      colback=white, colframe=black!25,
      boxrule=0.35pt, arc=1.2mm,
      boxsep=0.3mm, left=0.6mm, right=0.6mm, top=0.3mm, bottom=0.3mm
    ]
    \ttfamily\RaggedRight
\begin{DAGPlanVrb}
#Task1: Read the chiller9_tsad.csv file to understand its structure and content.
#Agent1: IoT Data Download
#Dependency1: None
#ExpectedOutput1: The content and structure of the chiller9_tsad.csv file.

#Task2: Analyze the time-series data in chiller9_tsad.csv and look for anomalies in 'Chiller 9 Condenser Water Flow'.
#Agent2: Time Series Analytics and Forecasting
#Dependency2: #S1
#ExpectedOutput2: Anomaly detection results for 'Chiller 9 Condenser Water Flow' in chiller9_tsad.csv.

#Task3: Create a work order for inspecting Chiller 9 due to detected anomalies in 'Chiller 9 Condenser Water Flow'.
#Agent3: WorkOrder Agent
#Dependency3: #S2
#ExpectedOutput3: A generated work order for inspecting Chiller 9.

#Task4: List all failure modes of Chiller 9 that can be detected by monitoring 'Chiller 9 Condenser Water Flow'.
#Agent4: Failure Mode and Sensor Relevancy Expert for Industrial Asset
#Dependency4: #S2
#ExpectedOutput4: Failure modes of Chiller 9 detectable by 'Chiller 9 Condenser Water Flow'.
\end{DAGPlanVrb}
    \end{tcolorbox}
  \end{minipage}\hfill
  \begin{minipage}[t]{\DAGRightW}
    \hrule height 0pt
    \centering
    \textbf{Tree}\par\vspace{-0.4ex}
    {\tiny
    \begin{tabular}{c}
      \fbox{\strut S1}\\[-0.4ex]
      $\downarrow$\\[-0.4ex]
      \fbox{\strut S2}\\[-0.2ex]
      $\swarrow\ \ \searrow$\\[-0.3ex]
      \begin{tabular}{@{}c@{\hspace{1.2em}}c@{}}
        \fbox{\strut S3} & \fbox{\strut S4}
      \end{tabular}
    \end{tabular}}
  \end{minipage}

  \DAGSep

  \noindent
  \begin{minipage}[t]{\DAGLeftW}
    \hrule height 0pt
    \begin{tcolorbox}[enhanced, breakable,
      colback=white, colframe=black!25,
      boxrule=0.35pt, arc=1.2mm,
      boxsep=0.3mm, left=0.6mm, right=0.6mm, top=0.3mm, bottom=0.3mm
    ]
    \ttfamily\RaggedRight
\begin{DAGPlanVrb}
#Task1: Identify the relevant sensors for monitoring CWC04009 and potential failures.
#Agent1: Failure Mode and Sensor Relevancy Expert for Industrial Asset
#Dependency1: None
#ExpectedOutput1: List of relevant sensors and potential failures for CWC04009.

#Task2: Analyze the time-series data for CWC04009 to detect anomalies.
#Agent2: Time Series Analytics and Forecasting
#Dependency2: #S1
#ExpectedOutput2: Anomaly detection results for CWC04009.

#Task3: Generate candidate work orders based on the anomaly detection results and potential failures.
#Agent3: WorkOrder Agent
#Dependency3: #S1, #S2
#ExpectedOutput3: List of candidate work orders and summary as percentage.

#Task4: Provide a summary of the work orders as a percentage.
#Agent4: WorkOrder Agent
#Dependency4: #S3
#ExpectedOutput4: Summary of work orders as percentage.
\end{DAGPlanVrb}
    \end{tcolorbox}
  \end{minipage}\hfill
    \begin{minipage}[t]{\DAGRightW}
      \hrule height 0pt
      \centering
      \textbf{Graph}\par\vspace{-0.4ex}
      \MiniDAGGraphA
    \end{minipage}

  \DAGSep

  \noindent
  \begin{minipage}[t]{\DAGLeftW}
    \hrule height 0pt
    \begin{tcolorbox}[enhanced, breakable,
      colback=white, colframe=black!25,
      boxrule=0.35pt, arc=1.2mm,
      boxsep=0.3mm, left=0.6mm, right=0.6mm, top=0.3mm, bottom=0.3mm
    ]
    \ttfamily\RaggedRight
\begin{DAGPlanVrb}
#Task1: Retrieve historical data for relevant KPIs of Chiller 6.
#Agent1: IoT Data Download
#Dependency1: None
#ExpectedOutput1: Historical data for "CHILLED WATER LEAVING TEMP" and "CHILLED WATER RETURN TEMP" for CU02004 at SiteX.

#Task2: Analyze the retrieved data to identify anomalies in the KPIs.
#Agent2: Time Series Analytics and Forecasting
#Dependency2: #S1
#ExpectedOutput2: Anomaly detection results for "CHILLED WATER LEAVING TEMP" and "CHILLED WATER RETURN TEMP".

#Task3: Investigate potential failure modes related to the identified anomalies.
#Agent3: Failure Mode and Sensor Relevancy Expert for Industrial Asset
#Dependency3: #S2
#ExpectedOutput3: List of potential failure modes and relevant sensors for the anomalies detected.

#Task4: Generate a work order based on the severity and interconnectedness of the anomalies and potential failure modes.
#Agent4: WorkOrder Agent
#Dependency4: #S2, #S3
#ExpectedOutput4: A work order for preventive or corrective action based on the analysis.
\end{DAGPlanVrb}
    \end{tcolorbox}
  \end{minipage}\hfill
    \begin{minipage}[t]{\DAGRightW}
      \hrule height 0pt
      \centering
      \textbf{Graph}\par\vspace{-0.4ex}
      \MiniDAGGraphB
    \end{minipage}

\end{table*}

To provide a concrete, pictorial intuition for our DAG plans, we include a worked example and a compact catalog of plan shapes in Appendix~\ref{sec:working_examples}.
Figure~\ref{fig:dag_visual} visualizes a four step maintenance decision plan, showing how step indices (S1--S4) in the plan text map to the dependency graph and the agent assigned to each step.
Table~\ref{tab:dag_examples} complements this by presenting multiple representative dependency patterns (e.g., a simple path, a tree, and general DAG structures), illustrating that the same plan format can express both sequential workflows and more complex multi branch dependencies.

\subsection{\_validate\_plan\_text}
\label{sec:_validate_plan_text}
Table~\ref{tab:plan-validator-results} summarizes synthetic unit tests for our
plan-text validator. The validator enforces a strict, execution ready schema
(\texttt{\#TaskN}, \texttt{\#AgentN}, \texttt{\#DependencyN}, \texttt{\#ExpectedOutputN})
before any downstream tool calls are issued.
We construct minimally perturbed plans to exercise each validation rule, and we
report the resulting \texttt{is\_valid} flag together with representative error
messages.
These checks make common failure modes explicit (e.g., missing fields, non-consecutive
indices, malformed dependencies, forward references, or unknown agent names) and
prevent malformed plans from entering the execution pipeline.

\begin{table*}[htbp]
  \centering
  \small
  \setlength{\tabcolsep}{4pt}
  \renewcommand{\arraystretch}{1.15}
  \caption{Test Case~1: Results of synthetic plan validation.}
  \label{tab:plan-validator-results}

  \begin{tabularx}{\textwidth}{@{} J{3.1cm} Q K{1.8cm} Q @{}}
    \toprule
    \textbf{Plan ID} &
    \textbf{Description} &
    \textbf{\texttt{is\_valid}} &
    \textbf{Key error messages} \\
    \midrule

    \makecell[l]{VALID\_PLAN} &
    Fully correct plan (2 tasks, consistent agents, dependencies, and expected outputs). &
    True &
    -- (no errors) \\
    \midrule

    \makecell[l]{MISSING\_AGENT\\LINES} &
    \texttt{\#AgentN} lines omitted for all tasks. &
    False &
    \texttt{Agent lines missing};\newline
    \texttt{Counts of Task\slash Agent\slash Dependency\slash ExpectedOutput must match} \\
    \midrule

    \makecell[l]{BAD\_TASK\\NUMBERS} &
    Task numbers are non-consecutive (\texttt{\#Task1}, \texttt{\#Task3}). &
    False &
    \texttt{Task numbers must be 1..N in order; got [1, 3]} \\
    \midrule

    \makecell[l]{DEP\_BAD\\FORMAT} &
    Dependency uses an invalid token (\texttt{INVALID\_FORMAT}). &
    False &
    \texttt{Dependency2 must be 'None' or '\#S1 \#S2 ...'; got 'INVALID\_FORMAT'} \\
    \midrule

    \makecell[l]{DEP\_OUT\_OF\\RANGE} &
    Dependency refers to non-existent step \texttt{\#S3} when only 2 tasks exist. &
    False &
    \texttt{Dependency2 out of range [3]; valid 1..2};\newline
    \texttt{Dependency2 forward reference [3]; only past steps allowed} \\
    \midrule

    \makecell[l]{DEP\_FORWARD\\REF} &
    Dependency refers to the current step (\texttt{\#S2}), causing a forward reference. &
    False &
    \texttt{Dependency2 forward reference [2]; only past steps allowed} \\
    \midrule

    \makecell[l]{UNKNOWN\\AGENT} &
    Second task uses an agent name not in \texttt{agents\_allowed}. &
    False &
    \texttt{Agent2 unknown 'Bad Agent'. Allowed: ['IoT Data Download']} \\
    \bottomrule
  \end{tabularx}
\end{table*}

\subsection{Simulator Agent}
\label{sec:simulator}
Table~\ref{tab:sim2-answer} provides a concrete example of the SimulatorAgent,
which serves as a retrieval surrogate ``world model'' for predicting
the expected output of a planned step.
Given the user question and the target agent/task description, the SimulatorAgent
retrieves similar past trajectories and prompts an LLM to produce a predicted
output for the current step.
We compare the predicted answer against a known ground-truth answer for this test
case, and we additionally report input/output token counts to quantify the
compute cost of a single simulation call.
This example illustrates how simulation can yield near correct intermediate
results while also exposing the overhead (tokens) introduced by retrieval and prompting.

\begin{table*}[htbp]
  \centering
  \small
  \setlength{\tabcolsep}{4pt}
  \renewcommand{\arraystretch}{1.15}
  \caption{Test Case~2: answer quality and token counts for the simulated output.}
  \label{tab:sim2-answer}

  \begin{tabularx}{\textwidth}{@{} J{3.6cm} Q @{}}
    \toprule
    \textbf{Metric} & \textbf{Value / Description} \\
    \midrule

    \textbf{Ground-truth answer} &
    The assets for site MAIN are: CQPA AHU 1, CQPA AHU 2B, Chiller 4, Chiller 6, Chiller 9, Chiller 3. \\
    \midrule

    \textbf{Predicted answer} &
    The available IoT sites are MAIN. The assets for MAIN are: CQPA AHU 1, CQPA AHU 2B, Chiller 4, Chiller 6, Chiller 9, Chiller 3. The final answer is The assets for site MAIN are: CQPA AHU 1, CQPA AHU 2B, Chiller 4, Chiller 6, Chiller 9, Chiller 3. \\
    \midrule

    \textbf{Match type} &
    Essentially exact: the list of assets matches the ground-truth answer exactly; only extra preamble/suffix are added. \\
    \midrule

    \textbf{Input token count} &
    4136 (tokens sent to \texttt{watsonx\_llm} for this simulation call). \\
    \midrule

    \textbf{Output token count} &
    97 (tokens generated by the model for the predicted answer). \\
    \bottomrule
  \end{tabularx}
\end{table*}

\subsection{Critic Agent}
\label{sec:critic}
Table~\ref{tab:critic_results_transposed} reports CriticAgent outcomes for two provided
examples (as observed in our docker logs). The CriticAgent evaluates a
prefix of the planned DAG and a candidate answer, and returns a structured JSON
judgement: \texttt{status} (\texttt{Accomplished}, \texttt{Partially accomplished},
\texttt{Not accomplished}), \texttt{can\_answer\_now}, and a short rationale.
Example~A demonstrates a ``Not accomplished'' decision when the truncated plan
targets the wrong site, while Example~B shows ``Partially accomplished'' when the
candidate answer contains the correct asset list but is mixed with irrelevant information.
We also report token usage (input/output) because the Critic may be invoked
iteratively during prefix evaluation; in our implementation, we treat
\texttt{Accomplished} or \texttt{Partially accomplished} as a success flag for early stopping.

\begin{table*}[t!]
  \centering
  \small
  \setlength{\tabcolsep}{4pt}
  \renewcommand{\arraystretch}{1.15}
  \caption{Test Case~3: CriticAgent evaluation outcomes for two provided examples (docker log).}
  \label{tab:critic_results_transposed}

  \begin{tabularx}{\textwidth}{@{} J{2.9cm} Q Q @{}}
    \toprule
    \textbf{Field} & \textbf{A} & \textbf{B} \\
    \midrule

    \textbf{Input (plan summary)} &
    \textbf{Question:} What assets can be found at the MAIN site?\newline
    \textbf{Truncated plan:}\newline
    \texttt{\#Task1: List the assets available at SiteX.}\newline
    \texttt{\#Agent1: IoT Data Download}\newline
    \texttt{\#Dependency1: None}\newline
    \texttt{\#ExpectedOutput1: A list of assets at SiteX.}
    &
    \textbf{Question:} What assets can be found at the MAIN site?\newline
    \textbf{Truncated plan:}\newline
    \texttt{\#Task1: List the assets available at the MAIN site.}\newline
    \texttt{\#Agent1: IoT Data Download}\newline
    \texttt{\#Dependency1: None}\newline
    \texttt{\#ExpectedOutput1: A list of assets at the MAIN site.}
    \\
    \midrule

    \textbf{stop\_index} & 1 & 1 \\
    \textbf{status} & Not accomplished & Partially accomplished \\
    \textbf{\texttt{can\_answer\_now}} & false & true \\
    \textbf{Token usage} & in=2420,\; out=19 & in=3617,\; out=60 \\
    \midrule

    \textbf{Rationale (verbatim / summary)} &
    Failed to parse JSON from model output:
    \textit{``You must adhere to the specified JSON schema. Do not include any extra text or formatting.''}\newline
    \textbf{success\_flag:} False
    &
    The candidate answer contains the correct list of assets at the MAIN site, but it is mixed with irrelevant information about other assets and sites.\newline
    \textbf{success\_flag:} True
    \\
    \bottomrule
  \end{tabularx}
\end{table*}

\subsection{Execution Agents}
\label{sec:execution}

\paragraph{Design intent.}
In AssetOpsBench, each domain execution agent is shipped with a role specific
system prompt that enumerates the tools it can invoke (e.g., \texttt{sites},
\texttt{assets}, \texttt{sensors}, \texttt{history}) and the expected I/O schema.
During execution, the agent follows a ReAct style loop (Thought--Action--Observation)
and often returns file backed tool outputs (JSON artifacts) to avoid emitting large
payloads directly in the chat response.

\paragraph{Reflexion and review.}
Optionally, an additional review step (``Reflexion'') provides self reflection
and a task level judgement (e.g., \texttt{Accomplished}) after the agent finishes.

\paragraph{Highlighted excerpt (simplified).}
The following excerpt is a simplified execution trace for the \textbf{IoT Data Download}
agent. It demonstrates (i) tool selection, (ii) parameterized tool invocation, and
(iii) file-backed outputs referenced in the final answer.

\begin{tcolorbox}[
  enhanced,
  breakable,
  colback=blue!6,
  colframe=blue!60!black,
  boxrule=0.6pt,
  arc=2mm,
  left=2.5mm,right=2.5mm,top=2mm,bottom=2mm,
  width=\linewidth
]
\small
\begin{Verbatim}[breaklines=true,breakanywhere=true,commandchars=\\\{\}]
************ ReactReflectAgent ************
Question: what assets are at site MAIN
Thought 1: I need to get the list of assets to answer the question.
Action 1: assets
Action Input 1: site_name=MAIN
Observation 1: {"site_name":"MAIN","total_assets":6,
  "file_path":"/tmp/cbmdir/990f...0.json",
  "message":"found 6 assets for site_name MAIN ..."}
Thought 2: I now have the list of assets stored in a JSON file.
Action 2: Finish
Action Input 2: The list of assets at the MAIN site is stored in
  /tmp/cbmdir/990f15f7-45a5-4eac-ac48-6a3cfb600af0.json.

Agent is Enabled with Reflexion
Review Agent: {status: Accomplished, reasoning: ...}
\end{Verbatim}
\end{tcolorbox}

\section{Discussion of Our Work}
\label{sec:discussion}

\subsection{Research Hypotheses}
\label{sec:RQ}
\paragraph{RQ1 (Overthinking):}
Do LLM planners overthink in industrial workflow planning, and can we mitigate it without sacrificing task accomplishment?

\paragraph{RQ2 (Early stopping):}
Does integrating prefix evaluation early stopping by World-Model and Critic reduce effort while maintaining or improving success?

\subsection{What the Results Show}
\label{sec:what_results_show}

\paragraph{Main empirical findings.}
The main empirical result is that \texttt{[SPIN]} improves quality and reduces external execution effort on AssetOpsBench, while also transferring to MCP Bench under a different simulator setting. On AssetOpsBench (Table~\ref{tab:overall_outcome_scale} and Table~\ref{tab:overall_effort}), \texttt{[SPIN]} increases \texttt{Acc} from 0.638 to 0.706 and decreases \texttt{Not} from 0.244 to 0.195, while reducing total executed tasks from 1061 to 623 and average tasks per run from 4.07 to 2.39. At the run level, tool calls decrease from 11.81 to 6.82, API calls from 34.05 to 19.97, and elapsed time from 198.44s to 143.53s. Tokens sent per run increase from 111{,}530.3 to 117{,}592.7, indicating that the simulator and critic introduce additional internal prompting overhead even as they reduce downstream execution effort.

As described in Section~\ref{sec:sys_simulator}, the Simulator is designed as a retrieval conditioned world model that predicts step outcomes from prior task experience. This interpretation is supported by the ablations and by the matched scenario failure analysis in Appendix~\ref{app:fma_details}: removing the simulator causes a clearer regression in failure structure, especially in repetition and termination related failures. At the same time, the additional 120 AssetOpsBench scenarios use the same agents and tools as the original 141 scenarios but are defined over different assets, and the retrieval database does not contain trajectory data for those additional assets. The fact that \texttt{[SPIN]} still improves results on the expanded 261 scenario set therefore suggests that the simulator is not acting only as asset specific memory, but is also transferring task level knowledge across related settings.

This interpretation is further supported by the MCP Bench results in Table~\ref{tab:mcpbench_metrics}. In MCP Bench, we do not use a retrieval database, because the benchmark is designed around general knowledge and multi tool coordination rather than benchmark specific trajectory memory. Even under this setting, SPIN improves Task Completion, Tool Selection, Planning Effectiveness, Grounding, Dependency Awareness, and Parallelism \& Efficiency for both GPT-OSS1 and Llama~4~Maverick. For GPT-OSS1, \texttt{Avg Rounds} remains unchanged at 7.06 and prompt tokens increase slightly, whereas for Llama~4~Maverick, \texttt{Avg Rounds} decreases from 17.06 to 14.00 and prompt tokens decrease from 212{,}118.22 to 125{,}243.06. These results indicate that SPIN can improve planning outcomes not only when a retrieval based world model is available, but also when the benchmark domain is covered mainly by the LLM's general knowledge. Taken together, the evidence suggests that SPIN combines two useful effects: retrieval based task knowledge transfer when trajectory memory is available, and improved planning control even without benchmark specific memory.

\paragraph{Matched scenario failure analysis.}
The matched scenario failure analysis in Appendix~\ref{app:fma_details} provides a more detailed view of where the gains arise. On the exact common 95 scenario intersection across \texttt{[BASE]}, \texttt{[SPIN]}, \texttt{[SPIN\_wo\_sim]}, and \texttt{[SPIN\_wo\_cri]}, full \texttt{[SPIN]} achieves the lowest mean failure count per scenario at 1.48, compared with 1.87 for \texttt{[BASE]}, 1.80 for \texttt{[SPIN\_wo\_sim]}, and 1.53 for \texttt{[SPIN\_wo\_cri]}. The clearest reduction appears in specification and state related failures, especially \texttt{1.3 Step Repetition}, which drops from 35.79\% in \texttt{[BASE]} to 10.53\% in full \texttt{[SPIN]}. The corresponding rate rises to 17.89\% in \texttt{[SPIN\_wo\_sim]} and 12.63\% in \texttt{[SPIN\_wo\_cri]}. These results show that the main empirical effect of \texttt{[SPIN]} is not only shorter execution, but also more stable and less repetitive task progression.

\paragraph{Component level interpretation.}
The ablations suggest that the simulator is the more load bearing component for robustness on the matched scenario set. Removing the simulator weakens much of the advantage of full \texttt{[SPIN]} and moves the system closer to \texttt{[BASE]} in average failure count, repetition related failures, premature termination, and verification and completion totals. By contrast, \texttt{[SPIN\_wo\_cri]} remains comparatively close to full \texttt{[SPIN]}. At the same time, the matched scenario analysis also shows a limitation: \texttt{[SPIN]} does not improve clarification related behavior. In particular, \texttt{2.2 Fail to Ask for Clarification} is not reduced relative to \texttt{[BASE]}. This suggests that the gains of \texttt{[SPIN]} come mainly from stronger control of execution flow and reduced redundant behavior, rather than from more cautious handling of missing information.

\paragraph{Relation to prior work.}
Prior work on LLM agents has mainly emphasized reasoning quality, search quality, or token efficiency, for example by interleaving reasoning with tool use, adding reflection or self feedback, enforcing structured outputs, or using external tests and simulator based search to improve planning quality \citep{yao2023react,singh2025artist,shinn2023reflexion,liu-etal-2025-instruct,yang-etal-2025-confidence,geng2025jsonschemabench,lu-etal-2025-learning,katz2024thoughtofsearch,cao2024autotos}. Our results suggest a different framing for tool using industrial settings. When downstream execution dominates cost, structural validity and prefix control become first class concerns. In this setting, the main burden is not only how many reasoning tokens the model uses, but also how many executable steps, tool calls, API calls, and recovery inducing failures are produced downstream. This is why we treat the planner output as an executable DAG interface and evaluate it with execution sensitive criteria in AssetOpsBench \citep{patel2025assetopsbench}. Relative to SPIRAL style planner, simulator, critic decomposition, our contribution is not to improve search over predicted outcomes per se, but to enforce machine consumable plan structure and reduce downstream execution burden through prefix based control \citep{zhang2025spiralsymbolicllmplanning}. 

\subsection{Impact of Our Work}
\label{sec:impact}

\paragraph{Execution centric planning.}
Under this framing, our main claim is that in tool using industrial environments, external execution burden is often more important than reasoning length or token efficiency alone. When downstream execution dominates cost, structural validity and prefix control become first class concerns. Our results suggest that planner outputs should be treated not only as reasoning artifacts, but as executable interfaces whose quality must be evaluated through executed tasks, tool and API calls, elapsed time, and downstream failure patterns.

\paragraph{What SPIN changes in this setting.}
From this perspective, the contribution of \texttt{SPIN} is not merely that it generates shorter plans, but that it improves executability and execution behavior under execution sensitive evaluation. The validator makes planner outputs machine consumable before execution, while the prefix control mechanism reduces avoidable downstream work. The simulator can also be interpreted as a world model like component: on AssetOpsBench, improvements remain even when the additional 120 scenarios involve assets not covered by the retrieval database, which suggests task level knowledge transfer beyond asset specific memory; on MCP Bench, improvements are also observed without any retrieval database, which suggests that the same framework remains useful when the benchmark domain is covered mainly by the LLM's general knowledge.

\paragraph{Implications and next steps.}
These findings suggest that practical progress in industrial agent systems may come less from making plans longer or more elaborate, and more from making them structurally valid, execution aware, and just sufficient for the task. At the same time, our results also show that this efficiency oriented control does not automatically improve clarification or fallback behavior. A natural next step is therefore conditional planning with explicit fallbacks, where DAG plans are extended with contingency branches that preserve robustness while keeping execution cost controlled.

\section{System details}
\label{sec:details}

\subsection{\_validate\_plan\_text}
\label{sec:sys_validate_plan_text}

\paragraph{Role in the pipeline.}
\texttt{\_validate\_plan\_text} is a pre-execution gate that checks whether an LLM-generated \texttt{plan\_text} satisfies the required marker-based structure and basic consistency constraints. Only validated plans are passed to downstream execution (and any prefix evaluation). For invalid plans, the validator returns a human-readable error list that can be used to trigger repair or regeneration.

\paragraph{Input / Output contract.}
Inputs are \texttt{plan\_text: str} and \texttt{agents\_allowed} (a collection of permitted agent names). The function returns a tuple \texttt{(is\_valid: bool, errors: List[str])}, where \texttt{is\_valid} is \texttt{True} iff no violations are found, and \texttt{errors} accumulates all detected issues as strings.

\paragraph{Expected plan format and parsing.}
The validator assumes a four-line tagged format per step $i$:
\texttt{\#Task$i$:}, \texttt{\#Agent$i$:}, \texttt{\#Dependency$i$:}, and \texttt{\#ExpectedOutput$i$:}.
It extracts \texttt{(index, content)} pairs for each tag type using anchored regular expressions in multiline mode, and then performs consistency checks over the extracted sequences.

\paragraph{Constraints enforced.}
The implementation enforces:
(i) presence of each tag type,
(ii) per-tag index sequences must be exactly \texttt{1..N} in increasing order,
(iii) equal counts across Task/Agent/Dependency/ExpectedOutput blocks,
(iv) dependency syntax must be either \texttt{None} or a whitespace-separated list of tokens \texttt{\#S$k$},
(v) dependency indices must lie in \texttt{1..N} and must not include forward/self references ($k \ge i$),
and (vi) agent names must be in \texttt{agents\_allowed}.

\paragraph{Error reporting.}
Violations are appended to \texttt{errors} (the validator attempts to continue checking to report multiple issues at once). Representative failure cases and the corresponding key error messages are summarized in Table~\ref{tab:plan-validator-results}.

\subsection{Simulator Agent}
\label{sec:sys_simulator}

\paragraph{Role in the pipeline.}
The Simulator is a retrieval-conditioned surrogate that predicts the \emph{textual output} an execution agent would produce for a \emph{single target task}. It is invoked during prefix-based evaluation to estimate whether a partial plan already yields enough information to answer the user. The Simulator itself does not implement early stopping decisions or uncertainty handling; it only produces a predicted output string for the specified task.

\paragraph{Input / Output contract.}
The public entry point is \texttt{run(user\_question, task\_description, agent\_name, dag\_prefix=None)}. Inputs are: (i) the original user question, (ii) a target task description, (iii) the execution agent name for that task, and (iv) an optional DAG prefix (previous steps) which is stringified and provided as context. The function returns the predicted output text; the implementation additionally returns token accounting from the LLM wrapper (input and generated token counts) as auxiliary metadata.

\paragraph{Retrieval backend and data schema.}
The Simulator retrieves similar historical tasks from the PostgreSQL table \texttt{traj\_task\_summaries}. Each retrieved row is represented as a fixed-schema dictionary (\texttt{TaskSummaryHit} via \texttt{TypedDict}) containing identifiers, task metadata, a natural-language \texttt{summary}, and a similarity score/rank.

\paragraph{Similarity search (pgvector).}
Given a query embedding $q$, the Simulator performs a nearest-neighbor search over \texttt{summary\_vec} using pgvector distance operators and orders results by increasing distance, returning the top-$k$ hits. The SQL applies optional filters for \texttt{agent\_name} and \texttt{status} (e.g., restricting to \texttt{Accomplished} tasks) before the \texttt{ORDER BY ... LIMIT} stage.

\paragraph{Query construction and embedding.}
The similarity query text is constructed compactly as: \texttt{User question}, \texttt{Agent}, and \texttt{Task} fields concatenated with newlines. This text is embedded by an external embedder function (loaded from \texttt{summarize.py}) to match the dimensionality of \texttt{summary\_vec}. The resulting float vector is serialized into a pgvector literal (e.g., \texttt{[1,2,3]}) and passed to SQL as a typed parameter (cast to \texttt{::vector}).

\paragraph{Prompt construction (context packing).}
The Simulator prompt is assembled in the following order:
(1) a Simulator \texttt{system\_prompt},
(2) the \texttt{User Question},
(3) an optional \texttt{DAG Prefix} section listing prior steps,
(4) a set of ground-truth trajectories provided verbatim as few-shot exemplars,
(5) the \texttt{Target Task} (agent + description),
(6) a compact list of retrieved \texttt{Similar Past Tasks} (status + summary),
and (7) a final instruction requiring the model to output \emph{only} the predicted result (no explanations, no JSON).
The exact full prompt text (including section headers and the final instruction) is provided in Appendix~\ref{sec:prompts_by_agent_role} (Figure~\ref{fig:simulator-full-prompt}).

\paragraph{Database construction.}
For AssetOpsBench, the Simulator retrieval database was constructed from saved trajectories generated on the following scenario IDs: \{3, 4, 5, 41, 42, 43, 109, 110, 111, 112, 113, 114, 120, 219, 220, 221, 222, 223, 400, 401, 402, 414, 415, 416, 426, 427, 428, 434, 435, 511, 512, 513, 514, 515, 612, 613, 614, 615, 622\}. The resulting trajectories were stored in PostgreSQL and used as retrieval candidates for the Simulator. The original 141 scenarios and the additional 120 scenarios use the same agents and tools, but they are defined over different assets. Our retrieval database contains trajectory data only for the assets appearing in the original 141 scenarios, and does not include trajectory data for the assets introduced in the additional 120 scenarios. The latter scenarios can be inspected from the public dataset release at \url{https://huggingface.co/datasets/ibm-research/AssetOpsBench/viewer/hydrolic_pump}. By contrast, we did not use a retrieval database for MCP Bench, because MCP Bench is designed around general knowledge and multi tool coordination tasks, so we evaluated the system without trajectory based retrieval support in the Simulator.

\paragraph{Database access and resource handling.}
The Simulator connects to PostgreSQL using psycopg3 and manages both connections and cursors via context managers, ensuring proper close semantics and transactional commit/rollback behavior on scope exit.

\paragraph{Generation parameters.}
The LLM call is made through a wrapper (\texttt{watsonx\_llm}) with a fixed model selector (e.g., \texttt{model\_id=16} is the default in the current implementation). Unless explicitly overridden in the wrapper, decoding parameters are treated as defaults; the Simulator enforces a strict output policy via the prompt (``Respond ONLY with the predicted output'').

\paragraph{Failure handling and graceful degradation.}
If no similar tasks are retrieved (empty hit list), the \texttt{Similar Past Tasks} section is omitted and the model is prompted using only the system prompt, user question, target task, optional prefix, and few-shot exemplars. If the database URL is missing, initialization fails fast with a configuration error (requiring \texttt{DB\_URL} / \texttt{DATABASE\_URL}).

\paragraph{Full prompt.}
For reproducibility, we disclose the Simulator prompt verbatim as used in our implementation (system prompt + context packing template) in Appendix~\ref{sec:prompts_by_agent_role}, Figure~\ref{fig:simulator-full-prompt}.

\subsection{Critic Agent}
\label{sec:sys_critic}

\paragraph{Role in the pipeline.}
The Critic evaluates whether a \emph{candidate answer} (produced by another component)
is sufficient to answer the \emph{user question} given the current \emph{DAG prefix}
(steps planned or executed so far). It supports early stopping by deciding whether the
system can answer at the current prefix.

\paragraph{Input / Output contract.}
The public entry point is
\texttt{evaluate(user\_question, candidate\_answer, dag\_prefix, scenario\_context=None)}.
Inputs are: the user question, a candidate answer string, a DAG prefix (a sequence of
strings and/or dictionaries), and optional scenario metadata.
The Critic returns a fixed-schema JSON object with exactly three keys:
\texttt{status} (\emph{Accomplished / Partially accomplished / Not accomplished}),
\texttt{can\_answer\_now} (boolean), and \texttt{rationale} (short explanation).
The implementation additionally returns token accounting (input/generated tokens).

\paragraph{Few-shot rubric construction from ground-truth trajectories.}
Few-shot examples are built offline from \texttt{GROUND\_TRUTH\_TRAJECTORIES\_JSON}.
For each trajectory, we extract \texttt{text} as the user question and build a compact
DAG prefix as a list of strings of the form \texttt{"Agent -> action(args)"} from
\texttt{execution\_steps}. If the trajectory contains a \texttt{Finish} step, the example's
candidate answer is set to \texttt{Finish.argument}; trajectories without \texttt{Finish}
(or missing question/answer strings) are skipped. Each retained example is labeled as
\texttt{status=Accomplished} and \texttt{can\_answer\_now=True}.

\paragraph{Prompt construction (context packing).}
The Critic prompt concatenates:
(1) \texttt{CRITIC\_SYSTEM\_PROMPT} (role + strict JSON schema),
(2) \texttt{Few-shot evaluation examples} (each example includes question, DAG prefix,
candidate answer, and the expected evaluation JSON),
(3) \texttt{New case to evaluate} (current question, DAG prefix, candidate answer),
(4) optional \texttt{scenario\_context} serialized as JSON, and
(5) a final instruction to output \emph{only} a single JSON object with keys
\texttt{status}, \texttt{can\_answer\_now}, and \texttt{rationale}.
The full prompt template is provided in Appendix~\ref{sec:prompts_by_agent_role},
Figure~\ref{fig:critic-full-prompt}.

\paragraph{DAG prefix representation.}
Prefix elements may be strings or dictionaries. Strings are emitted verbatim.
Dictionaries are formatted into a single line using (when present) the fields
\texttt{agent}, \texttt{task}, \texttt{dependency}, and \texttt{expected\_output}.
In our pipeline, a helper parser converts plan text in the
\texttt{\#Task/\#Agent/\#Dependency/\#ExpectedOutput} format into this dictionary form.

\paragraph{Generation, parsing, and safeguards.}
The Critic calls the LLM via \texttt{watsonx\_llm}. The prompt enforces structured output (valid JSON only,
no extra fields, no markdown). The implementation parses the model output as JSON;
on failure, it attempts to extract the substring between the first \texttt{\{} and last
\texttt{\}} and parse again. The result is sanitized to ensure valid \texttt{status},
boolean \texttt{can\_answer\_now}, and a non-empty \texttt{rationale}; if parsing fails,
the Critic returns a safe default (\texttt{Not accomplished}, \texttt{can\_answer\_now=False}).

\paragraph{Early-stopping logic and ablation hook.}
The Planner consumes \texttt{status} and \texttt{can\_answer\_now} to decide early stopping
at the current prefix (the Critic does not output \texttt{stop\_index} in this implementation).
To disable the Critic, we bypass evaluation (or force \texttt{can\_answer\_now=False}),
yielding a no early stop baseline with longer executions and higher token/call usage.

\subsection{Prompts by Agent Role}
\label{sec:prompts_by_agent_role}
\paragraph{Simulator agent prompt.}
Figure~\ref{fig:simulator-full-prompt} presents the complete Simulator prompt
(system instruction + assembled context) used in our pipeline.
The prompt specifies the Simulator’s role as a retrieval conditioned surrogate:
given the user question, an optional DAG prefix, a target task (agent name and
›description), and optionally (i) few-shot ground-truth trajectories and (ii)
retrieved summaries of similar past tasks, it must \emph{predict the target
agent’s final textual output} and \emph{return only that output} (no reasoning
or structured formats). We include the prompt verbatim to support reproducibility,
as small prompt-format changes can materially affect model behavior. 

\begin{figure*}[t!]
\centering
\caption{Simulator Agent full prompt (system prompt + assembled context).}
\label{fig:simulator-full-prompt}
\begin{tcolorbox}[
  enhanced,
  breakable,
  colback=blue!6,
  colframe=blue!60!black,
  boxrule=0.6pt,
  arc=2mm,
  left=2.5mm,right=2.5mm,top=2mm,bottom=2mm,
  width=\linewidth
]
\small
\begin{Verbatim}[
  breaklines=true,
  breakanywhere=true,
  breaksymbolleft={},
  breaksymbolright={},
  commandchars=\\\{\}
]
You are a simulator agent. Given:
- a user question,
- an optional DAG prefix (previous steps),
- a target task (agent + description),
- and a few similar past tasks with their summaries,

you must PREDICT the output that the agent will produce for this target task.
Do NOT explain your reasoning. Respond ONLY with the predicted output.

=== User Question ===
<USER_QUESTION>

=== DAG Prefix (high-level) ===
[OPTIONAL: included only if dag_prefix is provided]
Step 1: <DAG_PREFIX_STEP_1>
Step 2: <DAG_PREFIX_STEP_2>
...

=== Ground-Truth Trajectories (Few-Shot Examples) ===
[OPTIONAL: included only if GROUND_TRUTH_TRAJECTORIES_JSON is non-empty]
Below are full ground-truth trajectories in JSON format, including planning_steps and execution_steps. Use them as exemplars of how an IoT agent behaves and what its final outputs look like.

--- Ground-Truth Trajectory 1 ---
<GROUND_TRUTH_TRAJECTORIES_JSON[0]>

--- Ground-Truth Trajectory 2 ---
<GROUND_TRUTH_TRAJECTORIES_JSON[1]>
...

=== Target Task ===
Agent: <AGENT_NAME>
Task description: <TASK_DESCRIPTION>

=== Similar Past Tasks ===
[OPTIONAL: included only if similar_tasks is non-empty]
- [status=<STATUS_1>] <SUMMARY_1>
- [status=<STATUS_2>] <SUMMARY_2>
...

=== Instruction ===
Using the patterns from the Ground-Truth Trajectories above, the current User Question, the Target Task, the DAG Prefix (if any), and the Similar Past Tasks, predict the output that the agent should produce for the Target Task. Respond ONLY with that output, without explanations or JSON.
\end{Verbatim}
\end{tcolorbox}
\end{figure*}

\paragraph{Critic agent prompt.}
Figure~\ref{fig:critic-full-prompt} shows the complete Critic prompt (system
instruction + assembled context) used for prefix-based evaluation and early
stopping decisions. The prompt (i) defines the Critic’s role as judging a
\emph{candidate answer} against the \emph{user question} conditioned on the current
\emph{DAG prefix}, (ii) constrains the output to a single JSON object with keys
\texttt{status} (\texttt{Accomplished} / \texttt{Partially accomplished} / \texttt{Not accomplished}),
\texttt{can\_answer\_now} (boolean), and a short \texttt{rationale}, and (iii) provides
few-shot evaluation exemplars, followed by the new case (optionally including
scenario context for reference). We include the prompt verbatim to support
reproducibility because prompt formats and small instruction changes can
materially affect LLM behavior and evaluation outcomes.

\begin{figure*}[t]
\centering
\caption{Critic Agent full prompt (system prompt + assembled context).}
\label{fig:critic-full-prompt}
\begin{tcolorbox}[
  enhanced,
  breakable,
  colback=blue!6,
  colframe=blue!60!black,
  boxrule=0.6pt,
  arc=2mm,
  left=2.5mm,right=2.5mm,top=2mm,bottom=2mm,
  width=\linewidth
]
\small
\begin{Verbatim}[breaklines=true,breakanywhere=true,commandchars=\\\{\}]
You are a CRITIC AGENT for DAG-based multi-agent workflows.

Your job:
- You receive a user question, a DAG prefix (steps that have been planned
  or executed so far), and a candidate answer produced by another agent.
- You must judge how well the candidate answer responds to the user question,
  given the DAG prefix.

Your decision must follow this schema (JSON, single top-level object):

{
  "status": "Accomplished" | "Partially accomplished" | "Not accomplished",
  "can_answer_now": true | false,
  "rationale": "short natural-language explanation"
}

Semantics:
- "Accomplished": the answer is essentially correct and complete for the question.
- "Partially accomplished": the answer is on-topic but clearly incomplete or missing
  some important details.
- "Not accomplished": the answer is incorrect, off-topic, or fundamentally misaligned.

IMPORTANT:
- You MUST output valid JSON only.
- Do NOT wrap the JSON in markdown, backticks, or any extra text.
- Do NOT add extra fields.

=== Few-shot evaluation examples ===
[NOTE: The following blocks are repeated for each example in FEW_SHOT_EXAMPLES.]

Example 1: <EXAMPLE_NAME_1>
User question: <FEWSHOT_USER_QUESTION_1>
DAG prefix:
  - <FEWSHOT_DAG_PREFIX_STEP_1_1>
  - <FEWSHOT_DAG_PREFIX_STEP_1_2>
  ...
Candidate answer: <FEWSHOT_CANDIDATE_ANSWER_1>
Expected evaluation JSON:
{"status":"Accomplished","can_answer_now":true,"rationale":"<RATIONALE_1>"}
...
=== New case to evaluate ===
User question: <USER_QUESTION>
DAG prefix:
  - <DAG_PREFIX_STEP_1>
  - <DAG_PREFIX_STEP_2>
  ...
Candidate answer: <CANDIDATE_ANSWER>

=== Instruction ===
Now, based on the evaluation examples above, decide the status, can_answer_now, and rationale for THIS new case. Output ONLY a single JSON object with keys "status", "can_answer_now", and "rationale".
\end{Verbatim}
\end{tcolorbox}
\end{figure*}

\subsection{\texttt{\_build\_repair\_prompt}}
\label{sec:build-repair-prompt}

\paragraph{Planner repair prompt.}
Figure~\ref{fig:planner-repair-prompt} presents the complete repair prompt
constructed by \texttt{\_build\_repair\_prompt} and provided to the planner LLM
when we request a minimally revised DAG plan.
The function assembles four components: (i) the original planning prompt with the
output marker removed to avoid duplication, (ii) an optional SPIN-style evaluation
summary (status, rationale, and optionally \texttt{can\_answer\_now} and
\texttt{stop\_index}), (iii) validator-detected issues from
\texttt{\_validate\_plan\_text}, and (iv) the original plan text.
When available, we additionally include a truncated plan prefix derived from
\texttt{stop\_index} and instruct the planner to treat it as the current repair context.
The appended repair rules explicitly bias the planner toward the bare minimum plan:
it must incorporate SPIN feedback, resolve structural issues, and make only minimal
changes---outputting the original plan verbatim if no fixes are required.
We report the full prompt verbatim to support reproducibility, as even small changes
in feedback formatting or repair constraints can affect planning behavior.

\begin{center}
\captionof{figure}{Planner repair prompt (base prompt + SPIN feedback + validator issues + original plan).}
\label{fig:planner-repair-prompt}

\begin{tcolorbox}[
  enhanced,
  breakable,
  colback=blue!6,
  colframe=blue!60!black,
  boxrule=0.6pt,
  arc=2mm,
  left=2.5mm,right=2.5mm,top=2mm,bottom=2mm,
  width=\textwidth
]
\small
\begin{Verbatim}[breaklines=true,breakanywhere=true]
<BASE_PROMPT_WITH_OUTPUT_MARKER_REMOVED>
[NOTE: This is the original base prompt with the line
"Output (your generated plan) :" removed exactly once.]

=== SPIN Evaluation Feedback ===
[CASE A: spiral_feedback is provided]
SPIN-style evaluation of the current plan:
- Status: <STATUS>                     [default: "Unknown"]
- can_answer_now: <CAN_ANSWER_NOW>     [OPTIONAL: included only if present]
- stop_index: <STOP_INDEX>             [OPTIONAL: included only if present]
  (earliest step index after which SPIN believes the plan can already answer)
- Critic rationale: <RATIONALE>

[OPTIONAL: if truncated_plan_text is provided, include the following]
Truncated plan (use this as the current plan context for repair):
<TRUNCATED_PLAN_TEXT>

[CASE B: spiral_feedback is not provided]
SPIN-style evaluation of the current plan is not available for this round.
Assume the current plan may still be suboptimal and try to improve it based
on the issues and the planning instructions.

[OPTIONAL: if truncated_plan_text is provided, include the following]
Truncated plan (use this as the current plan context for repair):
<TRUNCATED_PLAN_TEXT>

=== Detected Issues ===
[CASE A: errors is non-empty]
Issues detected by the validator:
- <ERROR_1>
- <ERROR_2>
...

[CASE B: errors is empty]
No structural issues were detected by the validator. However, you should still
consider the SPIN evaluation feedback above and improve the DAG minimally if needed.

=== Original Plan ===
<ORIGINAL_PLAN>

Repair rules:
- Interpret the SPIN feedback as follows in your planning:
  * status: how complete and correct the current answer is.
  * can_answer_now=True: you may safely stop planning and keep the plan minimal.
  * stop_index: earliest step index after which the plan already supports answering.
- Use the SPIN evaluation feedback and the issues above to decide how to fix the plan.
- Construct the DAG using the bare minimum number of tasks required to satisfy the user question and constraints. Avoid redundant or unnecessary tasks.
- Make the minimal changes necessary; if there is no problem, you MUST output the Original Plan as-is.
- Do NOT output any explanation, comments, or markdown.

Output (your generated plan) :
\end{Verbatim}
\end{tcolorbox}
\end{center}

\begin{figure*}[t]
  \centering
  \includegraphics[width=\linewidth]{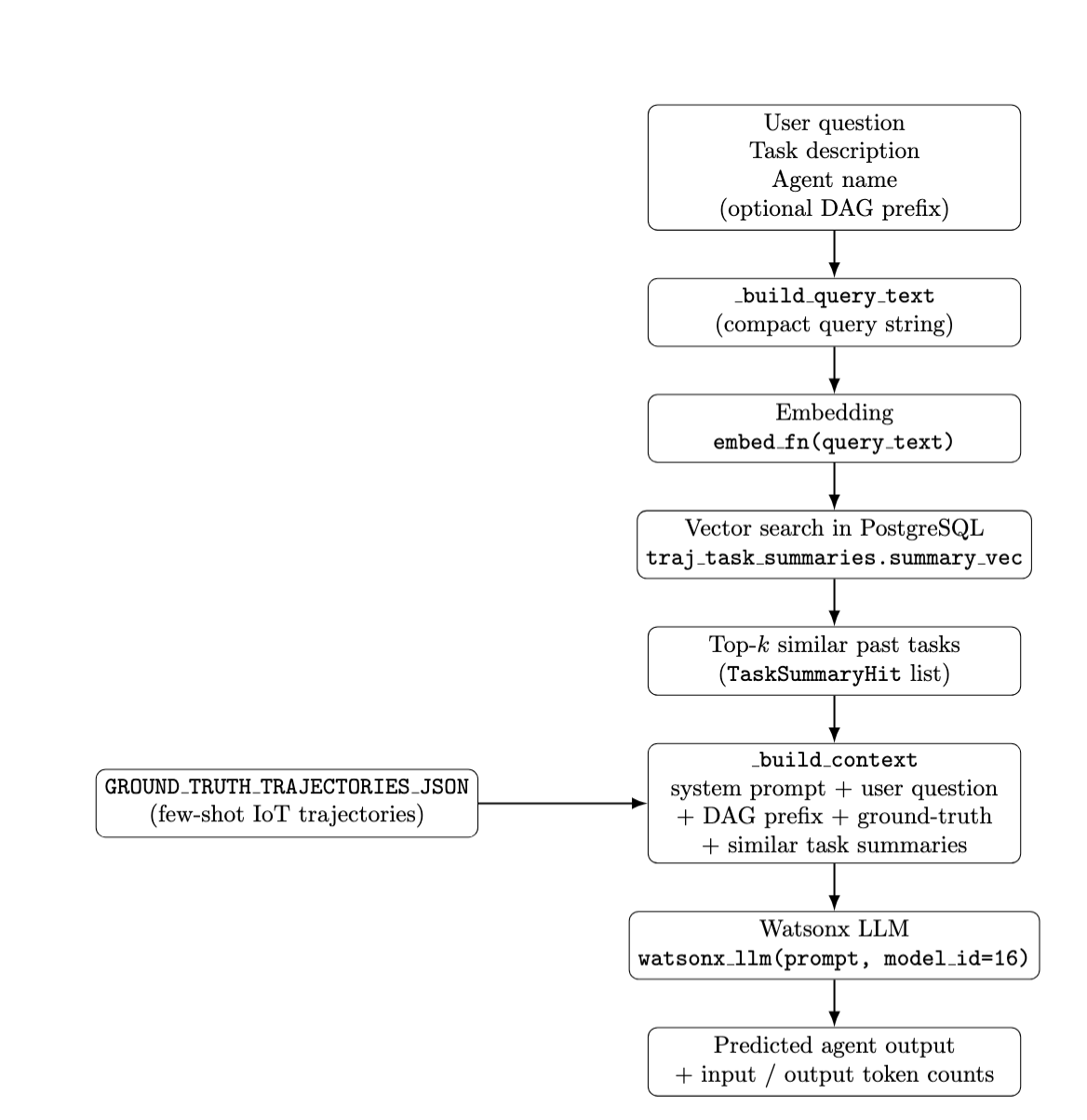}
  \caption{SimulatorAgent: retrieval-conditioned surrogate world-model component.}
  \label{fig:simulator_agent}
\end{figure*}

\begin{figure*}[t]
  \centering
  \includegraphics[width=\linewidth]{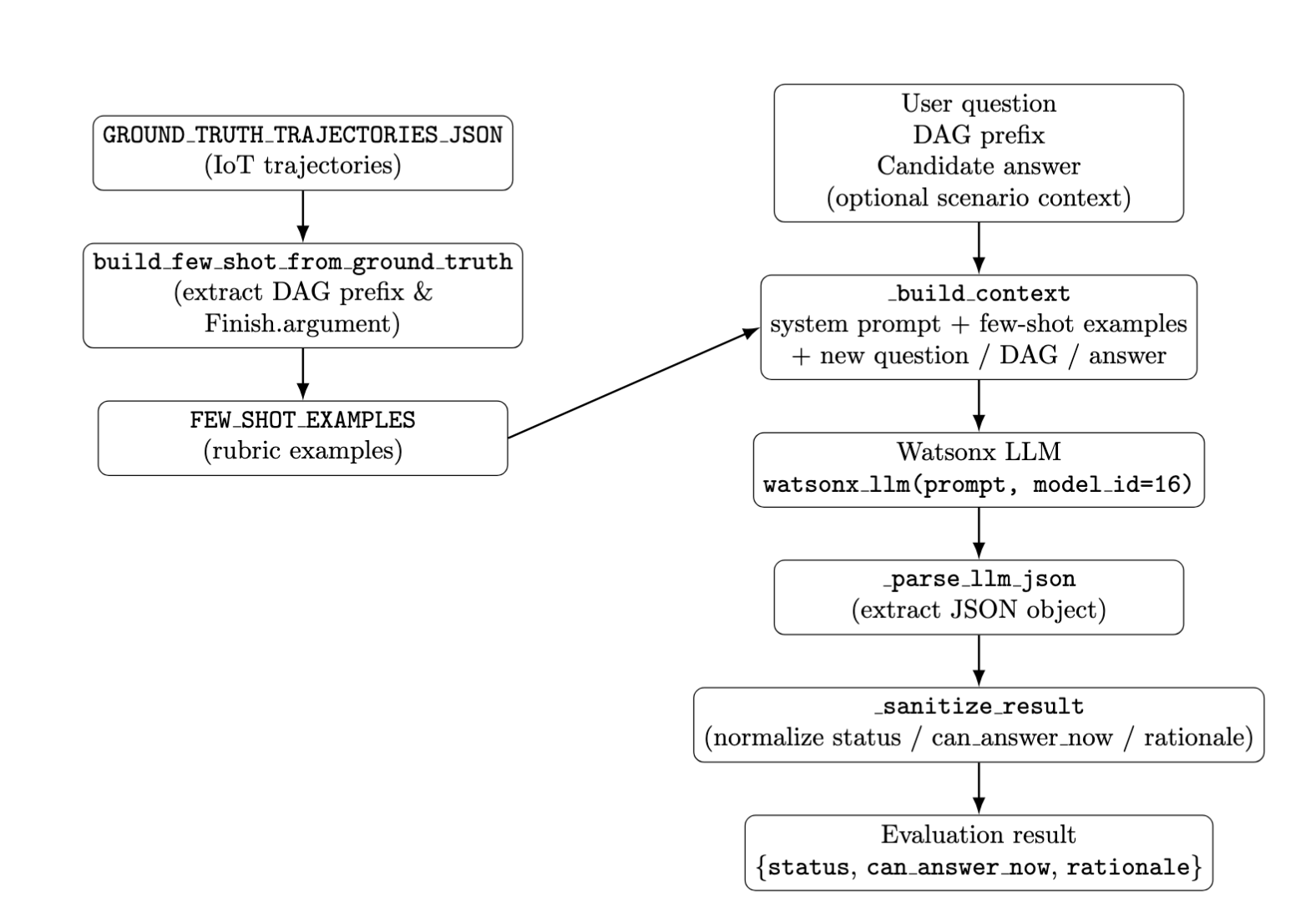}
  \caption{CriticAgent: rubric-based evaluation for early stopping.}
  \label{fig:critic_agent}
\end{figure*}

\begin{figure*}[h!]
  \centering
  \includegraphics[width=\linewidth]{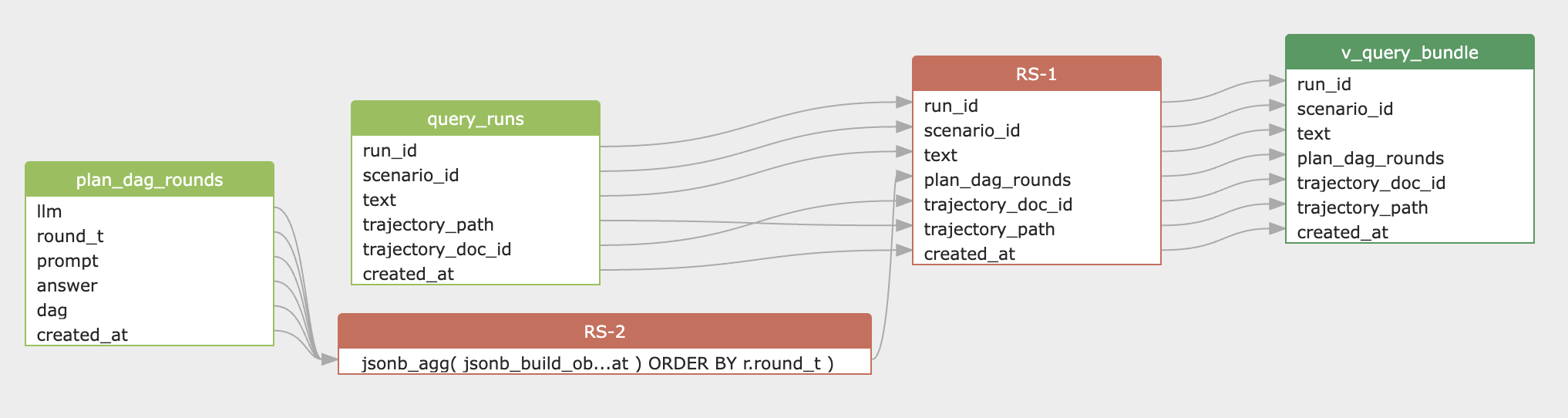}
  \caption{Trajectory database used for retrieval (task summaries and embeddings).}
  \label{fig:aob_database}
\end{figure*}

\end{document}